%% file: main.tex
\documentclass[acmsmall]{acmart}

\setcopyright{cc}
\setcctype{by}
\acmDOI{10.1145/3808171}
\acmYear{2026}
\acmJournal{PACMSE}
\acmVolume{3}
\acmNumber{FSE}
\acmArticle{FSE164}
\acmMonth{7}
\acmSubmissionID{fse26mainb-p1132-p}
\received{2026-02-19}
\received[accepted]{2026-03-24}

\usepackage[most]{tcolorbox}
\usepackage{xspace}
\usepackage{enumitem}
\usepackage[edges]{forest}
\usepackage{stfloats}
\usepackage{tabularray}
\usepackage{multirow}
\usepackage{mwe}
\usepackage{wrapfig}
\usepackage{colortbl}
\usepackage{siunitx}
\usepackage[linesnumbered,ruled,vlined]{algorithm2e}

\usepackage{amsmath}
\usepackage{pifont}

\usepackage{enumitem}
\usepackage{wrapfig}
\usepackage{hhline}

\usepackage{caption}
\usepackage{subcaption}
\usepackage[flushleft]{threeparttable}

\input{micro.tex}

\begin{document}

\title{Understanding, Detecting, and Repairing Real-World In-Context-Learning-Based Text-to-SQL Errors}

\author{Jiawei Shen}
\orcid{0009-0002-9349-5841}
\authornote{Both authors contributed equally to this research.}
\affiliation{%
  \institution{East China Normal University}
  \city{Shanghai}
  \country{China}
}
\email{51265902113@stu.ecnu.edu.cn}

\author{Chengcheng Wan}
\orcid{0000-0001-9162-9688}
\authornotemark[1]
\authornote{Chengcheng Wan and Geguang Pu are corresponding authors.}
\affiliation{%
  \institution{East China Normal University}
  \city{Shanghai}
  \country{China}
}
\affiliation{%
  \institution{Shanghai Innovation Institute}
  \city{Shanghai}
  \country{China}
}
\email{ccwan@sei.ecnu.edu.cn}

\author{Ruoyi Qiao}
\orcid{0009-0008-0373-2481}
\affiliation{%
  \institution{East China Normal University}
  \city{Shanghai}
  \country{China}
}
\email{10215101483@stu.ecnu.edu.cn}

\author{Jiazhen Zou}
\orcid{0009-0005-9180-258X}
\affiliation{%
  \institution{East China Normal University}
  \city{Shanghai}
  \country{China}
}
\email{51275902126@stu.ecnu.edu.cn}

\author{Hang Xu}
\orcid{0009-0002-2061-2285}
\affiliation{%
  \institution{East China Normal University}
  \city{Shanghai}
  \country{China}
}
\email{51275902085@stu.ecnu.edu.cn}

\author{Yuchen Shao}
\orcid{0009-0009-0414-7521}
\affiliation{%
  \institution{East China Normal University}
  \city{Shanghai}
  \country{China}
}
\email{51265902036@stu.ecnu.edu.cn}

\author{Yueling Zhang}
\orcid{0000-0002-2542-2619}
\affiliation{%
  \institution{East China Normal University}
  \city{Shanghai}
  \country{China}
}
\email{ylzhang@sei.ecnu.edu.cn}

\author{Weikai Miao}
\orcid{0009-0006-2038-439X}
\affiliation{%
  \institution{East China Normal University}
  \city{Shanghai}
  \country{China}
}
\email{wkmiao@sei.ecnu.edu.cn}

\author{Geguang Pu}
\orcid{0000-0001-9750-8334}
\authornotemark[2]
\affiliation{%
  \institution{East China Normal University}
  \city{Shanghai}
  \country{China}
}
\email{ggpu@sei.ecnu.edu.cn}

\renewcommand{\shortauthors}{J. Shen, C. Wan, R. Qiao, J. Zou, H. Xu, Y. Shao, Y. Zhang, W. Miao, and G. Pu}

\begin{abstract}
Large language models (LLMs) have been adopted for text-to-SQL tasks, utilizing their in-context learning (ICL) capability to translate natural language questions into SQL queries. However, such a technique faces correctness problems. In this paper, we conduct the first comprehensive study of text-to-SQL errors of ICL-based techniques. Our study covers four representative ICL-based techniques, five basic repairing methods, two benchmarks, and two LLM settings. We find that text-to-SQL errors are widespread and summarize 27 error types of 7 categories. We also find that existing repairing attempts have limited correctness improvement while having high computational overhead and many mis-repairs. Based on these findings, we propose \tool, a novel text-to-SQL error detection and repairing framework. The evaluation demonstrates that \tool outperforms existing solutions by repairing 13.8\% more queries with \edit{a negligible number of} mis-repairs and reducing 67.4\% \edit{repair latency}.  The artifact is publicly available at GitHub. 

\end{abstract}




\maketitle

\section{Introduction}

\input{sections/1-introduction.tex}

\section{Background}
\input{sections/2-background.tex}

\section{Understanding Text-to-SQL Errors}
\input{sections/3-Understanding.tex}

\section{Effectiveness of Existing Repairing Solutions}
\input{sections/4-Effectiveness.tex}

\section{\tool: Detecting and Mitigating Text-to-SQL Errors}
\input{sections/5-Tool.tex}

\section{Threats to Validity}
\input{sections/6-Threats.tex}

\section{Related Work}
\input{sections/7-related_work.tex}

\section{Conclusion}
\input{sections/8-conclusion.tex}

\section*{Data Availability}
The artifact is publicly available at GitHub~\cite{artifact}.
% It is also uploaded as the supplementary material.

\begin{acks}
We thank all anonymous reviewers for their valuable feedback. This paper is supported by the National Natural Science Foundation of China (Grant No. 62402183, 92582108, 62372181), the Shanghai Special Program for Promoting High-Quality Industrial Development (Project No. 250668, 250203), the CCF-Huawei Populus Grove Fund (Grant CCF-HuaweiTC202304) and the Shanghai Driverless Train Control System Engineering Technology Research Center (No. SUTC-2024KT-03).
\end{acks}

\bibliographystyle{ACM-Reference-Format}
\bibliography{citation.bib}

\end{document}

%% file: micro.tex
\newcommand{\edit}[1]{#1}

\newcommand{\others}{\text{17}\xspace}

\newcommand{\tool}{\textsc{MapleDoctor}\xspace}
\newcommand{\code}[1]{\texttt{#1}}

\newcommand{\bird}{\textsc{Bird}\xspace}
\newcommand{\spider}{\textsc{Spider}\xspace}
\newcommand{\sciencebenchmark}{\textsc{ScienceBenchmark}\xspace}
\newcommand{\scibench}{\shortstack{\textsc{Science}\\\textsc{Benchmark}}\xspace}

\definecolor{todocolor}{rgb}{0.9,0.1,0.1}
\definecolor{donecolor}{rgb}{0.1,0.6,0.4}

\newcommand{\eg}{\hbox{\emph{e.g.}}\xspace}
\newcommand{\ie}{\hbox{\emph{i.e.}}\xspace}
\usepackage{pythonhighlight}

\definecolor{codegreen}{rgb}{0,0.6,0}
\definecolor{codegray}{rgb}{0.5,0.5,0.5}
\definecolor{codepurple}{rgb}{0.58,0,0.82}
\definecolor{backcolour}{rgb}{0.97,0.97,0.95}
\definecolor{forestgreen}{rgb}{0.28,0.62,0.37}

\lstdefinestyle{mystyle}{
    backgroundcolor=\color{backcolour},   
    commentstyle=\color{codegray},
    keywordstyle=\color{codepurple},
    numberstyle=\tiny\color{codegray},
    stringstyle=\color{blue},
    basicstyle=\ttfamily\footnotesize,
    breakatwhitespace=false,         
    breaklines=true,                 
    captionpos=b,                    
    keepspaces=true,                 
    numbers=left,                    
    numbersep=5pt,                  
    showspaces=false,                
    showstringspaces=false,
    showtabs=false,                  
    tabsize=4,
}

\lstdefinestyle{SQL}{
    language = SQL,
    backgroundcolor=\color{backcolour},     % 背景色
    basicstyle = \small\ttfamily,           % 基本样式 + 小号字体
    rulesepcolor= \color{gray},             % 代码块边框颜色
    breaklines = true,                  % 代码过长则换行
    numbers = left,                     % 行号在左侧显示
    numbersep=5pt,                  
    showspaces=false,                
    showstringspaces=false,
    showtabs=false,                  
    tabsize=4,
    numberstyle = \tiny\color{codegray},               % 行号字体
    keywordstyle = \color{blue},            % 关键字颜色
    commentstyle =\color{codegray},        % 注释颜色
    showspaces = false,                 % 不显示空格
}

%% file: sections/1-introduction.tex
\subsection{Motivation}

\begin{wrapfigure}{r}{0.5\linewidth}
\centering
\includegraphics[width=\linewidth]{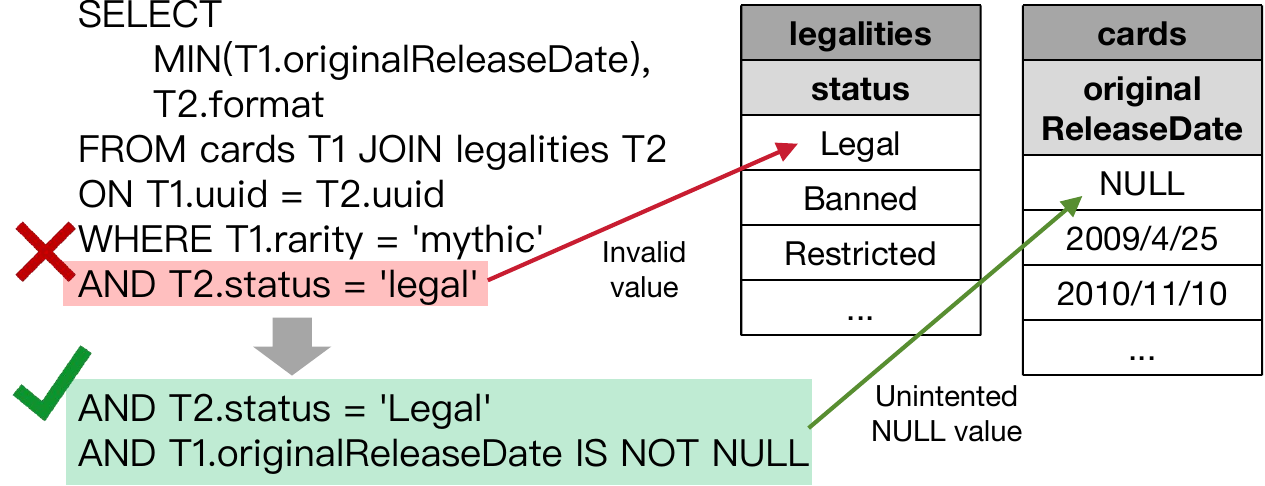}
\caption{An incorrect SQL query generated by DIN-SQL~\cite{din-sql}.}
\label{fig:intro-example}
\vspace{-0.2in}
\end{wrapfigure}

With the emergence of artificial intelligence~(AI), converting natural language~(NL) questions into structured query language~(SQL) has become a practical data management approach for non-expert users. Such text-to-SQL techniques allow users to interact with databases through conversation, eliminating the need of understanding database concepts and complex SQL syntax. Consequently, these techniques have been widely applied in data analysis~\cite{xue2023dbgpt,2023mirror,pandas}
, business intelligence~\cite{ibm_chatbi,baidu_chatbi,FinSQL}
, code repository management~\cite{DBLP:conf/issta/ZhangMKPL22}
, and other data-intensive domains, providing a user-friendly interface for accessing databases. Recently, in-context learning (ICL) \edit{-- a paradigm where models learn from a few demonstrations in the prompt without parameter updates --} of large language models (LLMs) has been adopted for text-to-SQL tasks~\cite{wang2024macsql,talaei2024chess,din-sql,xie-etal-2024-dea,gao2024dainsql,2023c3zeroshottexttosqlchatgpt,2024tasql,wang2024dacdecomposedautomationcorrection,shen-etal-2024-ASTRes,ACT-SQL,Distillery,CHASE-SQL,E-SQL,RSL-SQL}. However, they still face correctness problems and require repairing solutions. 

To better understand the correctness problem, consider an ICL-based text-to-SQL technique, DIN-SQL~\cite{din-sql}, in Figure~\ref{fig:intro-example}. Due to the hallucination and non-deterministic nature of LLMs, it generates a problematic SQL query for the question ``\emph{What is the release date and legal play format of the oldest mythic card?}''. It wrongly searches for an undefined value in an enumeration-style column \code{status}, violating the value specification. It also tries to find the minimum value of a column allowing \code{NULL} value. These errors are evident even without examining the question, highlighting the flaws of ICL-based techniques.

While recent work has proposed preliminary solutions for detecting and repairing SQL errors, they are often straightforward and have limited improvements. Several techniques~\cite{wang2024macsql,CHASE-SQL,talaei2024chess,E-SQL,RSL-SQL} categorize errors of their generation results according to SQL clauses and repair a small subset of these surface-level patterns. Others~\cite{din-sql,xie-etal-2024-dea} simply ask LLMs to validate and repair all the generated SQL queries, providing error descriptions or repairing examples. An earlier empirical study~\cite{ning2023errorstudy} studies SQL errors of non-LLM techniques, missing the unique errors introduced by LLMs (details in Section~\ref{sec:errors}). It also lacks error detection and repairing solutions.

Given these limitations, it is desirable to conduct an empirical study to understand the text-to-SQL errors of ICL-based techniques, based on which design a systematic, robust, and automated repairing solution. There exist several challenges:

1) \emph{How to obtain a high-quality error taxonomy that assists repairing?} While text-to-SQL errors could be easily classified by clause locations, such taxonomy does not help in understanding error root causes and lacks symptom information for error detection. In addition, the taxonomy should be unambiguous and cover as many errors as possible. How to design a helpful, comprehensive, and clear taxonomy is an open question.

2) \emph{How to avoid introducing new errors during repairing?} A practical error-repairing solution should minimize false alarms and avoid introducing new errors. Due to the generative nature of LLMs, we cannot fully rely on LLMs themselves to identify and resolve their own errors~\cite{huang2024largelanguagemodelsselfcorrect}. Instead, we should design a robust repairing solution with few false alarms, which is challenging.

3) \emph{How to detect and repair errors with low-overhead?} As a human-computer interaction scenario, high overhead significantly degrades the user experience. While invoking LLMs is a relatively effective solution, it is impractical to apply such expensive solutions to each SQL query.

\subsection{Contribution}

In this work, we conduct the first comprehensive study of text-to-SQL errors of ICL-based techniques and the effectiveness of their repairing methods. \

Through our study on four ICL-based techniques, two benchmarks, and two LLM settings, we construct a novel symptom-driven error taxonomy comprising 27 error types of 7 categories. Instead of error locations, it focuses on the ``edit path to correctness'', aiming to provide actionable solution for automated repair. We find that, 26.0\% of errors are \textit{intent-independent errors}, and 30.9\% to semantics, demonstrating a great improvement space (Section~\ref{sec:errors}).

We study 5 basic repairing methods from existing ICL-based techniques. We discover that these repairing attempts have limited correctness improvement --- they \edit{only fix 10.9-23.3\% (\ie, 75-161 out of 690) erroneous queries, at the cost of introducing 5.3-40.1\% (\ie, 4-55) more errors and 1.03-3.82$\times$ (\ie, 1.38-\SI{5.14}{s}) latency overhead.}

We also find that specifications and execution information are beneficial for LLM repairing (Section~\ref{sec:method_effectiveness}). 

Based on these findings, we propose \tool, a novel framework for efficiently detecting and repairing ICL-generated SQL queries. 
It identifies errors through symptom-based static analysis, which is built upon our error taxonomy.
Once an error is detected, \tool prioritizes rule-based repairs whenever a high-confidence fix exists; only when rules yield multiple plausible candidates or when execution exhibits suspicious symptoms indicative of semantic error, \tool invokes an LLM repairer with a constrained prompt.

Evaluation shows that \tool outperforms the state-of-the-art methods by repairing 13.8\% (\ie, 15) more erroneous SQL queries and reducing 84.9\%  (\ie, 45) mis-repairs, with 67.4\% less overhead (Section~\ref{sec:tool}).

%% file: sections/2-background.tex
\subsection{ICL-Based Text-to-SQL Techniques}

Text-to-SQL techniques aim to translate natural language questions into SQL queries in order to fetch data from the database. Given the effectiveness of LLMs, several ICL-based techniques are proposed, which provide data schema and other information in the prompt.
The ICL-based text-to-SQL workflow typically has three major stages. 

The \emph{pre-generation} stage creates condensed data schema information by only describing related tables and columns. Concrete column values related to the NL question are also included~\cite{talaei2024chess,CHASE-SQL,E-SQL}, for handling enumeration-style columns (\ie, ``Legal'' value in Figure~\ref{fig:intro-example}). 
ICL-based techniques also statically~\cite{wang2024macsql,talaei2024chess,gao2024dainsql,2023c3zeroshottexttosqlchatgpt,2024tasql,E-SQL,CHASE-SQL,Distillery} or dynamically~\cite{xie-etal-2024-dea, din-sql, wang2024dacdecomposedautomationcorrection, shen-etal-2024-ASTRes, ACT-SQL,luo2024PTD-SQL} provide several text-SQL pairs, serving as few-shot examples to enhance LLM capability on hard tasks.
Then, the \emph{generation} stage generates the SQL query with the prompt constructed with all information from the earlier stage. 
Finally, the \emph{post-generation} stage tries to detect and repair the potential errors in the generated query. Sometimes, it also examines consistency between multiple generated SQL queries~\cite{2023c3zeroshottexttosqlchatgpt, gao2024dainsql, li2024petsqlpromptenhancedtworoundrefinement, 2024mcssqlleveragingmultipleprompts}.

\subsection{Text-to-SQL Error Repairing}
\label{sec:background_repair}

Several existing ICL-based techniques propose repairing solutions. These solutions vary in error detection algorithms and supplementary information for LLM to understand and repair errors. 

\paragraph{Error Identification.}
Judging the correctness of SQL queries is inherently hard. While the compiler reports syntax errors, semantic errors are likely to escape.
One line of solutions, including DIN-SQL~\cite{din-sql}, CHESS~\cite{talaei2024chess} and DEA-SQL~\cite{xie-etal-2024-dea}, adopts chain-of-thought design and asks LLMs to detect errors for all generated queries by providing anti-pattern information. However, they have high computation costs and are likely to regard a correct query as wrong.
Another line of solutions applies simple rule-based solutions to identify error symptoms during query execution. For example, MAC-SQL~\cite{wang2024macsql} targets execution errors, empty results, and \code{NULL} values. These techniques only address a small subset of erroneous queries.

\paragraph{Supplementary Information.}
After identifying an error, the LLM is asked to re-generate the SQL query.
Several techniques use static prompts --- DIN-SQL simply asks for regeneration, and DEA-SQL provides examples of common error patterns.
In contrast, MAC-SQL provides database error messages and existence of empty results or \code{NULL} values. CHESS provides the query execution results. If the query-referenced value does not exist in the database, it also provides similar values in the referred column.
Without a thorough understanding of text-to-SQL errors, these techniques cannot provide precise and targeted information \edit{to improve performance}.

%% file: sections/3-Understanding.tex
\label{sec:errors}

\subsection{Methodology}

\subsubsection{Empirical Setup.}

There is a joint impact of prompt design and LLMs on the effectiveness of text-to-SQL techniques. Meanwhile, \edit{the prompt design of} these techniques are optimized specifically for the characteristics of the benchmark and LLM. Therefore, simply applying a technique to a new LLM or benchmark is likely to produce distorted results. 
\edit{To capture real-world text-to-SQL errors, we study the errors made by existing SOTA text-to-SQL solutions rather than artificial errors produced by our solutions. This setup ensures the analyzed errors are real-world and reflect genuine limitations of SOTA solutions.} We focus on the LLM and benchmark configurations that are shared across text-to-SQL techniques, and evaluate the corresponding technique implementation \edit{to avoid introducing extra errors from suboptimal prompting}. 

\textbf{Benchmark.} We adopt the dev split of two representative text-to-SQL benchmarks, \spider~\cite{spider} and \bird~\cite{li2024bird}, \edit{which have been adopted by the majority of recent ICL-based text-to-SQL techniques.} \spider dev split offers 20 databases and 1034 text-SQL pairs. \bird is a larger benchmark with more challenging tasks, whose dev split has 11 databases and 1534 text-SQL pairs. It covers more question types and SQL clauses, and provides additional context of evidence and database descriptions. All experiments are conducted in SQLite~\cite{sqlite}, which is a widely-used database management system (DBMS) supported by most of text-to-SQL benchmarks.

\textbf{Subject ICL-based techniques.} We select subject techniques based on (1) publicly-validated performance; (2) variance of error detection and repairing solutions; and (3) availability of open-source implementations. We select 4 representative techniques: MAC-SQL~\cite{wang2024macsql}, DIN-SQL~\cite{din-sql}, CHESS~\cite{talaei2024chess}, and DEA-SQL~\cite{xie-etal-2024-dea}\edit{, which are the top-performing open-source solutions as of our study conducted}.
This section focuses on the generation step, and leaves the post-generation step to Section~\ref{sec:method_effectiveness}.
Note that, DEA-SQL is only evaluated on \spider, as it does not provide the version for \bird. 

\textbf{LLM selection.} As the prompt design is correlated with the LLM family, we adopt the GPT family supported by all the subject techniques. We choose GPT-3.5-Turbo-0125 (GPT-3.5) and GPT-4o-2024-05-13 (GPT-4o) as representatives. DIN-SQL and CHESS are only evaluated with GPT-3.5, due to their huge token consumption \edit{(\ie CHESS takes >\$2,500 on \bird with GPT-4o).}

\textbf{Metrics.} As one NL question may correspond to multiple equivalent SQL queries, we use \emph{execution match} (EM) to judge query correctness, rather than exact string match. EM regards a query as correct only when its result set is the same as that of the gold query (\ie ground-truth). We adopt the implementation of \bird benchmark.

\edit{The empirical study covers various benchmarks and techniques, as summarized in Table~\ref{tab:empirical exp setup}.}

\begin{table}
    \centering
    \caption{Empirical study settings.}
    \vspace{-0.15in}
    \label{tab:empirical exp setup}
    % \small
    \footnotesize
    \edit{\begin{tabular}{c|c|c}
        \hline
        Evaluated Dataset   & GPT-3.5 & GPT-4o       \\ \hline
        MAC-SQL~\cite{wang2024macsql}   & \bird, \spider  & \bird, \spider \\ 
        DIN-SQL~\cite{din-sql}   & \bird, \spider  & -            \\ 
        CHESS~\cite{talaei2024chess}     & \bird, \spider  & -            \\ 
        DEA-SQL~\cite{xie-etal-2024-dea}   & \spider         & \spider \\ \hline
    \end{tabular}}
\end{table}

\subsubsection{Error Analysis.}
\label{sec:emprical_setup_error}
While all errors could lead to execution failure or incorrect query results, we focus on the errors on the ``edit path to correctness'', which provides clear guidance for correcting the SQL query step by step, rather than completely rewriting the query.
That is, we aim to create a taxonomy for the error symptoms that each is associated with a concrete mistake, as those shown in Figure~\ref{fig:intro-example}. Such a taxonomy assists in understanding the root causes of text-to-SQL errors, providing guidance for repairing. The error analysis is conducted in two phases.

In the first phase, we built initial error taxonomies by executing all ICL-based techniques with GPT-3.5 on the more challenging \bird benchmark to capture a wide spectrum of error types. We labeled error categories based on symptoms using an open card sorting approach~\cite{spencer2004card}. 
Specifically, in each iteration, 100 incorrect queries were randomly selected, and three co-authors independently labeled their error categories. They then cross-validated and discussed the labels until consensus was reached. Such an iteration was repeated until all 2460 incorrect SQL queries were analyzed. 
% 835+759+866=2460 {MAC-SQL, DIN-SQL, CHESS} in \bird using GPT-3.5

In the second phase, we used the initial taxonomies to label SQL queries from the rest of the settings. When an error could not be adequately described by existing taxonomy, or is hard to recognize, three co-authors convened to discuss it. If a new category was added to the taxonomy, all labeled errors were re-labeled with the new taxonomy.
This manual analysis process consumed approximately 840 person-hours.

\subsection{Error Taxonomy}

\begin{figure*}
\centering 
\includegraphics[width=\textwidth]{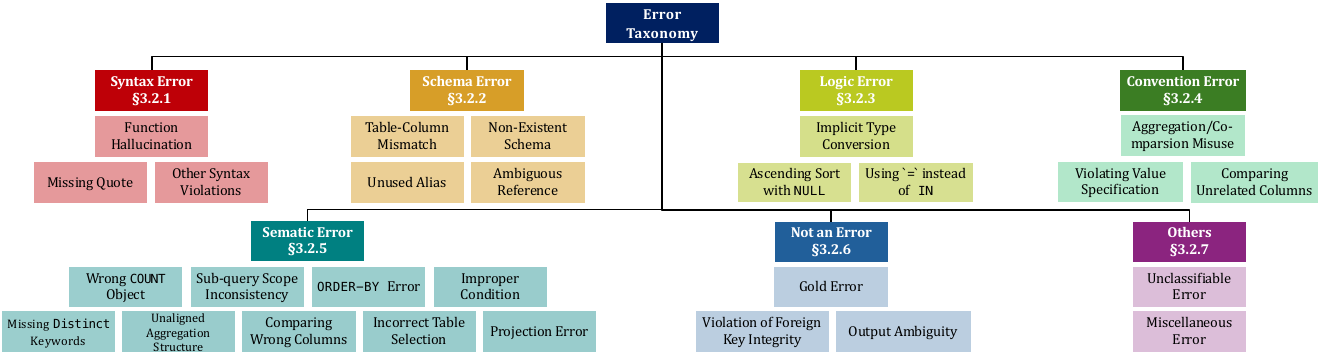}
\vspace{-0.25in}
\caption{Text-to-SQL error taxonomy for ICL-based techniques.}
\vspace{-0.1in}
\label{fig:error_taxonomy}
\end{figure*}

After manually studying 4602 incorrect SQL queries generated by ICL-based techniques, we build a two-level error taxonomy with 27 types, as illustrated in ~\autoref{fig:error_taxonomy}.

\subsubsection{Syntax Error.} The SQL query could not be parsed into a valid abstract syntax tree (AST), and thus fail to execute. We observe three types of syntax errors.  

\textbf{A1: Function Hallucination.} The generated SQL query invokes a function which seems semantically correct but does not exist in SQL standard or DBMS. All the observed hallucinated functions are designed for arithmetic and date processing. For example, CHESS wrongly invokes a non-existent function \code{DIVIDE(a, b)}, which should be operator \code{/}. As another example, even told to be executed in SQLite, CHESS still invokes \code{YEAR(x)} which is only implemented by MySQL. In fact, it should invoke \code{STRFTIME('\%Y', x)} instead.

\textbf{A2: Missing Quote.} The delimiters (\eg, single/double quotes, backtick and square brackets) do not properly enclose certain text strings and identifiers. It typically occurs on 
identifiers that contain spaces, special characters, or SQL reserved keywords. The \code{order} table
of \bird benchmark is an example, whose name would be regarded as a keyword unless quoted.

% Misc
\textbf{A3: Other Syntax Violations.} All other syntax errors, including unbalanced parentheses, incorrect clause ordering, missing necessary keywords, and others.

\subsubsection{Schema Error.} The SQL query is successfully parsed into an AST, but fails in schema resolution stage and thus cannot execute. We observe 4 types of schema errors. 

\textbf{B1: Non-Existent Schema.} 
The SQL query references a table or column whose name does not exist in database. It has two subcategories: (1) \emph{spelling error}, \eg, omitting characters; and (2) \emph{schema hallucination}, where LLM generates a semantically plausible but non-existent table/column name. 

\begin{wrapfigure}{r}{0.45\linewidth}
    \centering
    \vspace{-0.1in}
    \includegraphics[width=\linewidth]{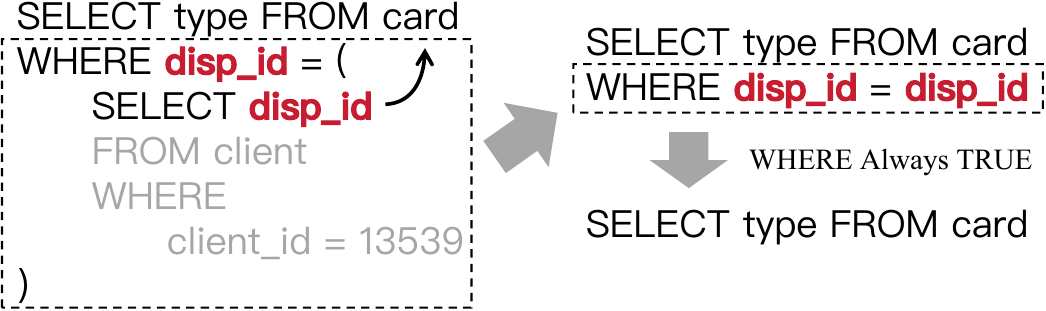}
    \vspace{-0.2in}
    \caption{
        Non-existent schema 
        \footnotesize{\textnormal{(\#161@\bird in MAC-SQL with GPT-3.5)}}.
    }
    \vspace{-0.2in}
    \label{fig:correlated_subquery}
\end{wrapfigure}

Surprisingly, not all non-existent schema errors trigger compiler error. 
For the non-existent table or column names in quotes, SQLite interprets them as string values rather than references. 
For the correlated sub-queries (\ie, an inner-level sub-query that references columns of an outer one), SQLite attempts to match the non-existent column to the table in outer query, which fails in most cases and causes schema errors. 
Figure~\ref{fig:correlated_subquery} is an example, which uses \code{disp\_id} column of \code{client} table to link two tables. Although \code{disp\_id} does not exist in that table, SQLite matches it with the \code{card} table in the outer level, making the outer-level \code{WHERE} clause always true.
Since these two scenarios involve SQLite's misinterpretation of non-existent columns, we also put them into this category.

\textbf{B2: Table-Column Mismatch.} The SQL query references a column that does not exist in the specified table of query, but appears in other tables of the database. It appears in two scenarios: (1) \emph{explicit mismatch}, where the column is explicitly associated with a table (\ie \code{SELECT T1.C1 FROM T1}), via name or alias; and (2) \emph{implicit mismatch}, where the query specifies the column without table information (\ie \code{SELECT C1 FROM T1}).

% 原 Alias Not Use
\textbf{B3: Unused Alias.} A table or a column is assigned an alias, but the references still use its original name, which is no longer valid. 

% 原 Qualifier Not Specified
\textbf{B4: Ambiguous Reference.} A query references a column that exists in multiple selected tables, but does not explicitly specify the corresponding table. For example, the referred column \code{C1} exists in both table \code{T1} and \code{T2}, but the generated SQL query is \code{SELECT C1 FROM T1, T2 ...}, without evidence of the exact table.

\subsubsection{Logic Error.} The SQL query passes schema resolution but contains logic errors which are obvious even unaware of the database and NL question. We observe 3 types of logic errors.

\begin{figure}[t]
\centering
\begin{minipage}{0.48\linewidth}
    \centering
    \includegraphics[width=0.7\linewidth]{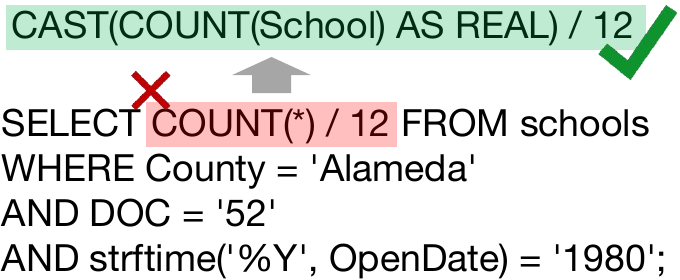}
    \vspace{-0.1in}
    \caption{
        Implicit type conversion and its fix \footnotesize{\textnormal{(\#47@\bird in CHESS with GPT-3.5)}}.
    }
    \label{fig:Implicit_type_convertion}
\end{minipage}\hfill
\begin{minipage}{0.48\linewidth}
    \centering
    \includegraphics[width=0.85\linewidth]{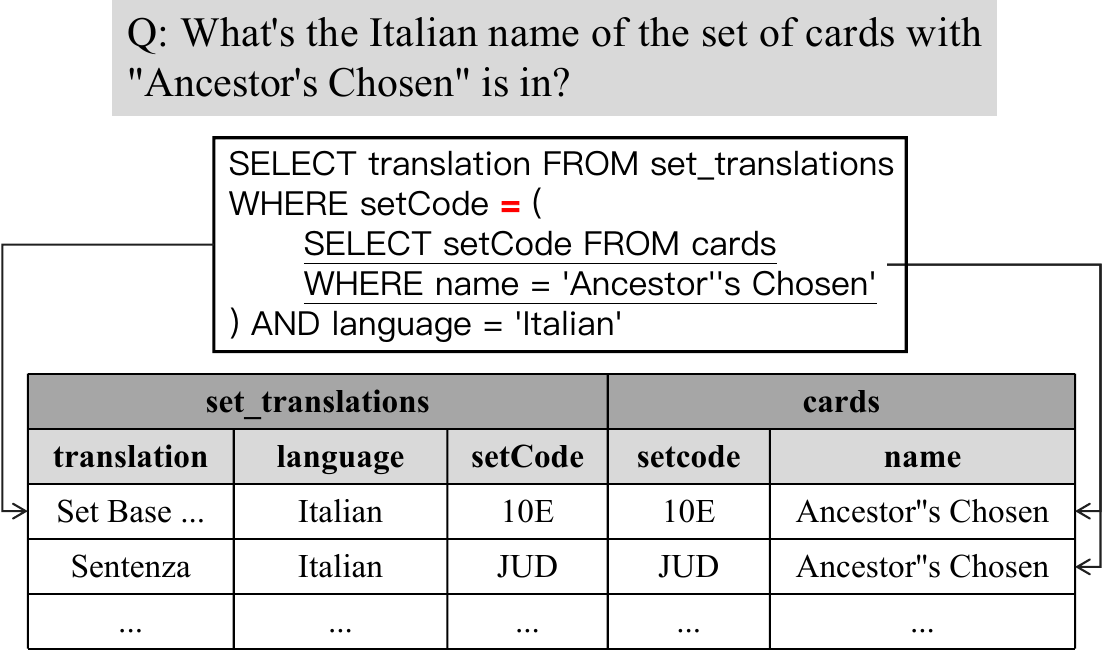}
    \vspace{-0.1in}
    \caption{
        Using = instead of IN \footnotesize{\textnormal{(\#462@\bird in MAC-SQL with GPT-4o)}}.
    }
    \vspace{-0.2in}
    \label{fig:IN_Eq}
\end{minipage}
\end{figure}

\textbf{C1: Implicit Type Conversion.} 
Database implicitly converts data to an incompatible type, which typically happens on multi-variate operators and aggregate functions. Such automatic conversion mechanism may lead to unexpected results. 
Figure~\ref{fig:Implicit_type_convertion} shows a concrete example. The generated query performs the division operation between two integer numbers, while it should explicitly turn one of them into float type to ensure a float result. In this example, the correct response is 1.416, while the SQL query outputs a floored value of 1.

\textbf{C2: Using = instead of IN.} 
The \code{=} operator is performed between a single value and a set of values, which actually should be \code{IN} operator. In such a case, SQLite will silently take the first element of the set and make the comparison, leading to a stricter condition. 
As shown in Figure~\ref{fig:IN_Eq}, while the sub-query returns two values, only the first one (\ie, \code{'10E'}) becomes the operand of the \code{=} operator. Therefore, the main query fails to, but should, retrieve the data entries with value \code{'JUD'}.

\textbf{C3: Ascending Sort with NULL.} 
When sorting a column containing \code{NULL} values, the latter are typically regarded as the smallest value, which may lead to an unexpected \code{NULL} output when a concrete number is expected.  
This error is common with the ascending sort and \code{MIN()} function in SQLite. A straightforward fix is to remove \code{NULL} values before invoking these operations.

\subsubsection{Convention Error.} The SQL query passes schema resolution but contains errors that can be identified just by analyzing database schema, without understanding of NL question. We observe 3 types of convention errors. 

\textbf{D1: Violating Value Specification.} 
In a column with value specifications, the SQL query attempts to find a value that violates the specification. It typically occurs on an enumeration-type or enumeration-style column. It also occurs on column with strict format, \eg, date and time. This type of error makes the query condition unsatisfiable, resulting in an empty query result. It has two typical causes: (1) \emph{wrong value}: the query examines the correct column but expects an impossible value; and (2) \emph{wrong column}: a wrong column is examined, making the expected values meaningless.
For example, in Figure~\ref{fig:intro-example}, the generated query attempts to find \code{'legal'} value in the \code{legalities.status} column. However, valid values for this column are \code{'Legal'}, \code{'Banned'}, and \code{'Restricted'}. Therefore, the condition in the \code{WHERE} clause will never be satisfied.

\textbf{D2: Aggregation/Comparison Misuse}
The SQL query applies numerical aggregation functions (\eg, \code{SUM} and \code{AVG}), ordering aggregation functions (\eg, \code{MAX} and \code{MIN}), and comparison operators (\eg, \code{<}, \code{=}, \code{>} and \code{ORDER BY}) on an improper column. 

For unique ID columns and enumeration-style columns using integer formats, applying numerical or ordering aggregation functions is inappropriate.
Similarly, or a non-numerical column whose comparison lacks practical significance (\eg, free text), it is also unreasonable to use a numerical aggregation function. Moreover, it often leads to incorrect type conversions. If the conversion fails, the system often defaults to treating the value as zero, causing misleading results.

\begin{figure}[t]
\centering
\begin{minipage}{0.48\linewidth}
    \centering
    \includegraphics[width=\linewidth]{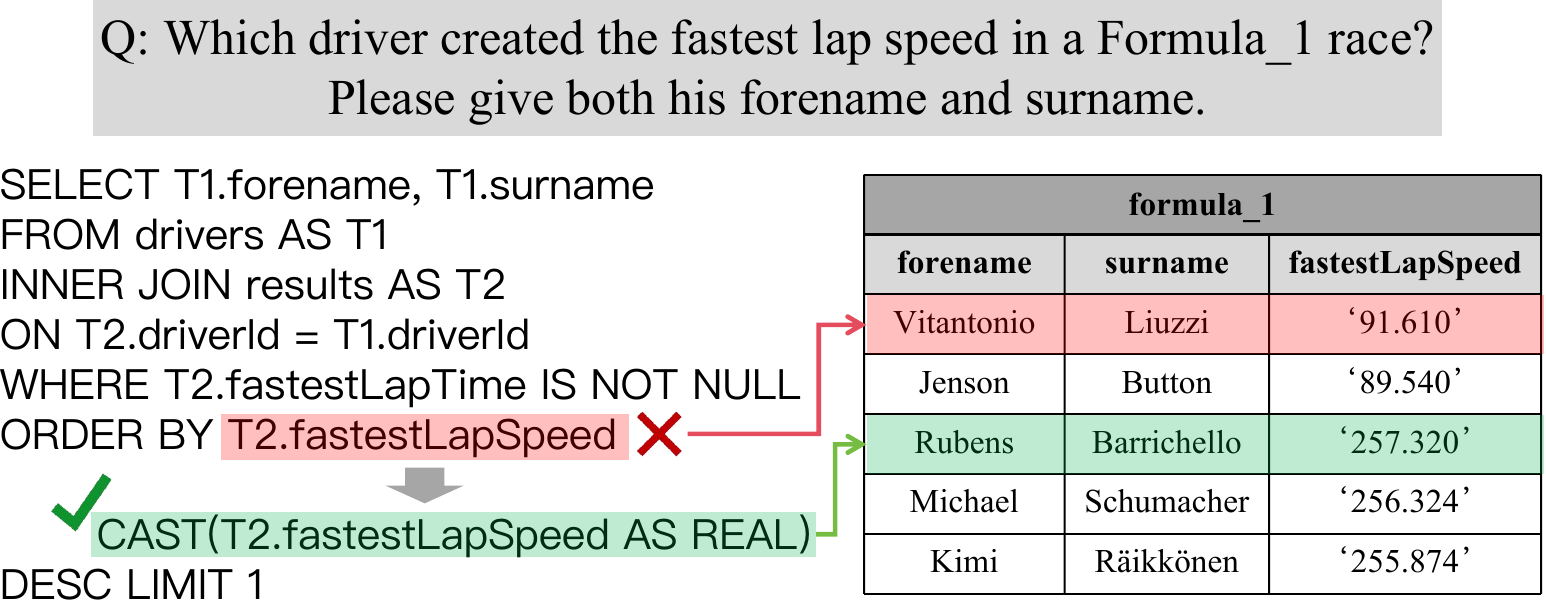}
    \vspace{-0.3in}
    \caption{
        Comparison misuse and its fix \footnotesize{\textnormal{(\#927@\bird, incorrect ground-truth)}}. 
        %, gold error 
    }
    \label{fig:Comparison_misuse}
\end{minipage}\hfill
\begin{minipage}{0.48\linewidth}
    \centering
    \includegraphics[width=\linewidth]{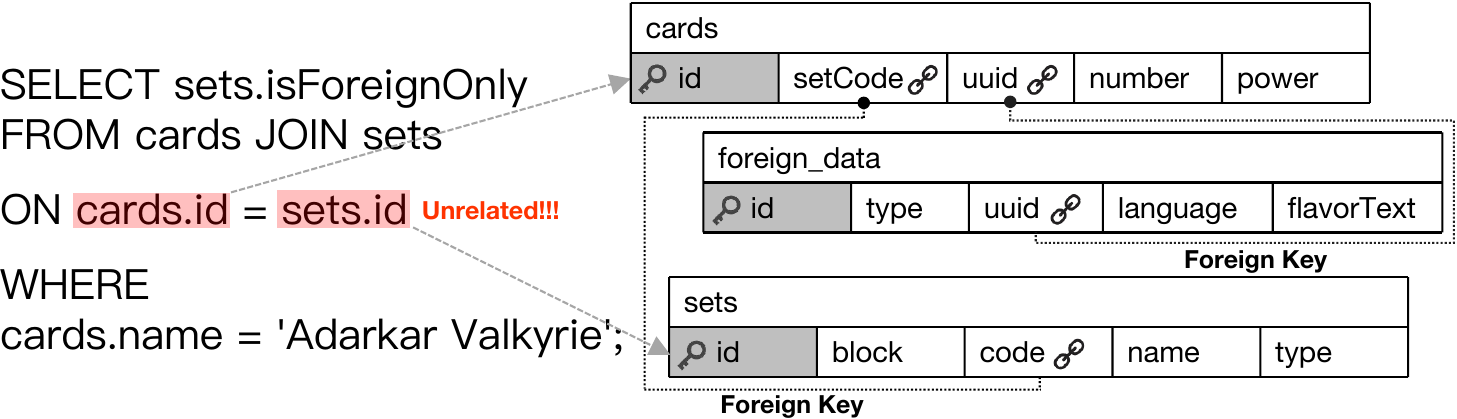}
    \caption{
        Comparing unrelated columns \footnotesize{\textnormal{(\#473@\bird in CHESS with GPT-3.5)}}.
    }
    \label{fig:OutputFormat_ComparingUnrelatedColumns}
\end{minipage}
\vspace{-0.2in}
\end{figure}

Particularly, for a text column, applying a comparison operator triggers string-based comparisons, which may yield unexpected outcomes. Figure~\ref{fig:Comparison_misuse} attempts to find the fastest driver. However, the text-type \code{fastestLapSpeed} column uses lexicographical ordering when finding the max value. Therefore, \code{'91.610'} is wrongly regarded as larger than \code{'257.320'}. This error could be fixed by a simple type conversion to ensure correct comparison.

\textbf{D3: Comparing Unrelated Columns.} 
The LLM mistakenly selects two unrelated columns and applies \code{IN}, \code{ON}, or comparison operators. 
Such error is common in the \code{JOIN} clause generated by LLMs, leading to either an empty set or a joined table with irrational relationship between columns.

Figure~\ref{fig:OutputFormat_ComparingUnrelatedColumns} shows a concrete example, where the \code{cards.id} and \code{sets.id} columns are wrongly selected in the \code{ON} condition of a \code{JOIN} clause. 
While these columns have the same names, they actually are unrelated: the former is the unique identifier for each card, and the latter is for each set. The correct columns for the \code{ON} condition are \code{cards.setCode} and \code{sets.code}.

\subsubsection{Semantic Error.} The SQL query passes schema resolution but contains semantic errors that could only be identified with the knowledge of database schema and the NL question.

\textbf{E1: Incorrect Table Selection.} 
The generated SQL query selects inappropriate tables. It has three primary forms: (1) a required table is omitted; (2) an unrequired table is included; and (3) a wrong table is selected in place of the correct one.

\textbf{E2: Projection Error.} 
The \code{SELECT} statement contains wrong columns or formats. For example, when answering ``\emph{Whose post has the highest popularity?}'', the SQL query erroneously selects \code{MAX(ViewCount)}, which only retrieves the highest popularity value. However, the question actually asks for names, which should be obtained from \code{DisplayName}.

\textbf{E3: Improper Condition.} The \code{WHERE} clause contains incorrect conditions, including incompleteness, redundancy, and other problems. 
Unlike the \emph{violating value specification} error, this type of error does not violate database schema or contain syntax errors. Instead, it is only semantically incorrect with respect to the NL question. 

\begin{wrapfigure}{r}{0.45\linewidth}
    \centering
    \includegraphics[width=\linewidth]{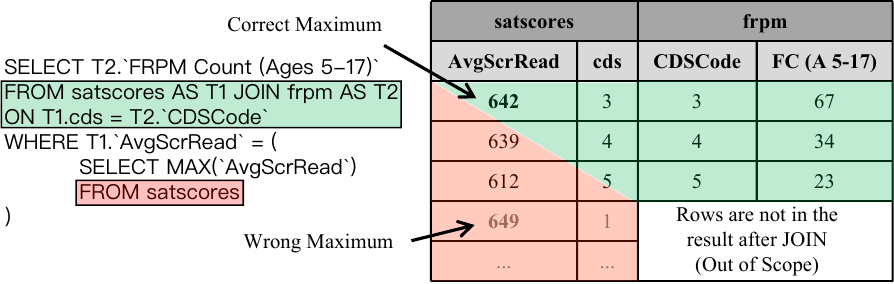}
    \vspace{-0.1in}
    \caption{
        Sub-query scope inconsistency \footnotesize{\textnormal{(\#10@\bird in DIN-SQL with GPT-3.5)}}. 
    }
    \label{fig:Subquery_Scope_Inconsistency}
\end{wrapfigure}

\textbf{E4: Sub-query Scope Inconsistency.} The scope of a sub-query does not align with the main query, leading to unexpected query results. 
Typically, the sub-query attempts to find a max/min value or to filter results, but only covers a different set of the data that the main query aims to retrieve.
Figure~\ref{fig:Subquery_Scope_Inconsistency} shows a concrete example, which aims to find students with the highest average reading score. 
The sub-query traverses the entire \code{satscores} table, while the main query only traverses the joined table, a subset of the former. Therefore, it is likely that the main query cannot find the maximum value that the sub-query identified, leading to an empty query result.

\textbf{E5: Unaligned Aggregation Structure.} 
The generated SQL query misses or has incorrect/redundant aggregation components, including aggregate functions, \code{GROUP BY} clause, and \code{HAVING} clause. These components affect how the query results are aggregated and expanded, which should match with the NL question.

\textbf{E6: Wrong COUNT Object.} 
The \code{COUNT} function is applied to an incorrect column or object. While LLMs understand the counting requirements of questions, they often struggle to identify the correct counting object. For example, when answering ``How many molecules have a triple bond type?'', they simply use \code{COUNT(*)} instead of counting the unique molecule IDs from the \code{bond} table.

\textbf{E7: ORDER-BY Error.} 
The \code{ORDER BY} clause has (1) wrong object, where the wrong column or expression is specified for sorting; or (2) wrong order, where the sorting direction
% (\ie, ascending and descending) 
is inverted.

\textbf{E8: Missing DISTINCT Keyword.} 
The generated query omits the \code{DISTINCT} keyword and thus retrieves redundant data entries. 
This error particularly happens in queries without aggregation function or the \code{LIMIT} keyword.

\textbf{E9: Comparing Wrong Columns.} The two columns, although related, are wrongly compared in a query.
It typically happens in the \code{JOIN} operations, where multiple columns are syntactically correct to join two tables while not all of them are semantically correct according to the NL question. 
For example, in the \code{financial} database of \bird, the query wrong uses the \code{district.district\_id} column to join \code{client} and \code{account} tables, as they all refer to it. However, their relationship is actually stored in the \code{disp} table.

\subsubsection{Not an Error.} The generated SQL query is regarded as incorrect due to the error of benchmark, ambiguity of NL question, or improper implementation of correctness judgment.

\textbf{F1: Gold Error.} The benchmark provides an incorrect ground-truth (gold query). The gold error is only caused by benchmark quality, regardless of the actual correctness of the generated query.

\textbf{F2: Violation of Foreign Key Integrity.} 
The database violates foreign key constraints, leading to unexpected query results. It is common in SQLite, which disables foreign key validation by default~\cite{sqlite3_fk_enforcement}.
The \code{toxicology} database of \bird is a concrete example. The \code{molecule.molecule\_id} column is the foreign key of \code{bond.molecule\_id} column. However, the latter contains values unrecorded in the former. Therefore, SQLite cannot correctly join these two tables.

\textbf{F3: Output Ambiguity.} Due to ambiguity of natural language, the NL questions may lack information to clearly specify the expected query output, including the required output format, the inclusion/exclusion of \code{NULL} values, and how to handle extremum-related questions where multiple entries that share the same extreme value.

\subsubsection{Others.} \label{sec:others_error}
We have tried our best to build a comprehensive taxonomy to guide the detection and fixing of text-to-SQL errors. However, more than $\frac{1}{3}$ erroneous SQL queries remain unclassified --- they have severe mistakes and require a complete rewrite to obtain correct query results. We cannot identify its edit path, and thus label them as \emph{unclassifiable errors}. Additionally, there are \others types of error that \edit{occur fewer than 6 times}, which are grouped into \emph{miscellaneous errors}.

\subsection{Errors of ICL-based Techniques}  

\input{tables/chapter_3_detailed_error_statistics}

With the taxonomy, we have categorized and summarized the errors in SQL queries generated by all four studied ICL-based text-to-SQL techniques, as shown in Table \ref{table: error result}. Around 15.4\% of all queries have \emph{not-an-error} problems, which indicates the unreliability of their correctness judgment.
Therefore, we omit these queries when discussing the error statistics. 

Overall, among all the 6136 generated SQL queries in \bird benchmark that have reliable correctness judgment, 47.8\% of them contain at least one true error. As the tasks in \spider dataset are simpler, only 26.8\% of the queries are erroneous. 
In many cases, an incorrect SQL query contains multiple errors belonging to different categories, which are labeled with all these error types.

% Syntax Error, Execution Failure and Logic Error
\subsubsection{Intent-Independent Errors.}\hfill

\emph{Intent-independent errors} are the ones can be identified without knowing the NL questions --- syntax, schema, logic, and convention errors. 
Around 26.0\% errors in \bird belong to this category, and 25.8\% in \spider. Such prevalence shows the inner flaws of ICL-based techniques in constructing correct SQL queries that are executable and seemingly feasible on a concrete database.

\begin{tcolorbox}[width=\linewidth, boxrule=0pt, sharp corners=all,
 left=2pt, right=2pt, top=0pt, bottom=0pt, colback=gray!20]
\textbf{Finding 1}: Intent-independent errors contribute 26.0\%(25.8\%) of all errors in \bird(\spider) dataset. There is much improvement space for ICL-based techniques to generate correct queries.
\end{tcolorbox}

Syntax errors constitute 18.8\% of all errors that cause execution failures. More than 40\% of syntax errors in \bird are \emph{function hallucination}, highlighting the LLM hallucination problem in function invocation. Different DBMSs have different standards and libraries. For example, \code{YEAR(x)} function is unsupported by SQLite, but valid in MySQL~\cite{mysql}. As LLMs do not distinguish DBMSs, this function is wrongly invoked in SQLite and fails to execute. In other words, when performing domain-specific tasks, the general LLMs suffer from negative impacts from out-of-domain knowledge. 

Meanwhile, \emph{missing quote} and \emph{other syntax violations} occur from time to time, due to LLMs' generative nature. Unlike rule-based solutions, LLMs do not guarantee strict-format responses~\cite{shao2024vortexrippletempiricalstudy}. ICL-based techniques inevitably generate queries with syntax errors.

Schema errors cause more than 80\% execution failures, indicating that current ICL-based techniques face challenges in comprehending database schema information.
More than 80\% of MAC-SQL's schema errors are \emph{table-column mismatch}.
Meanwhile, the \emph{unused alias} is only observed with CHESS. After thoroughly investigating these two techniques, we conclude that such error distribution is likely caused by the prompt design.

\begin{tcolorbox}[width=\linewidth, boxrule=0pt, sharp corners=all,
 left=2pt, right=2pt, top=0pt, bottom=0pt, colback=gray!20]
\textbf{Finding 2}: Syntax errors and schema errors cause 18.8\% and 81.2\% execution failures, respectively. They are caused by the LLM's inner flaws and prompt designs.
\end{tcolorbox}

Logic errors only constitute 2.1\% of the total errors. More than half of the logic errors are \emph{using = instead of IN}, where LLM wrongly assumes a sub-query always returns a single entry. We also observe that most (38 of 41) errors are with MAC-SQL, as it tends to decompose instructions instead of merging.
More than 90\% of the logic errors occur in \bird, as it involves more complex computation, on which \emph{implicit type conversion} typically happens. It also has databases with more entries and \code{NULL} values, as well as contains more nested queries.

Convention errors account for 13.4\% and 12.9\% of all errors in \bird and \spider, respectively.
\emph{Violating value specification} is the most common convention error (around 70\%), as LLM prompts cannot cover all the value specifications that maps NL keywords to a specific database value. 
It is also hard for LLMs to infer such specifications from the name of tables and columns. Most techniques either omit value specification, or randomly select some of them. Even when all relevant value specifications are provided, LLMs may still fail to follow them due to their generative nature.

Among the errors of \emph{comparing unrelated columns}, CHESS has $1-8\times$ more errors than other techniques. 
The reason is that CHESS mistakenly prunes out foreign key columns when constructing prompt, and thus the LLMs lack information to properly join tables. The CHESS authors have confirmed our reported bug and patched their implementation~\cite{chess_issue}.

\begin{tcolorbox}[width=\linewidth, boxrule=0pt, sharp corners=all, left=2pt, right=2pt, top=0pt, bottom=0pt, colback=gray!20]
\textbf{Finding 3}: Understanding database schema is a challenging task for ICL-based techniques. Around 12.6\% of text-to-SQL errors are logic errors and convention errors, among which \emph{violating value specification} is the most frequent error type.
\end{tcolorbox}

\subsubsection{Semantic Errors.}\hfill
\label{sec:errors_result_semantic}

Semantic errors account for 36.1\% and 17.7\% of all errors in \bird and \spider benchmark, respectively.
They are caused by misinterpretation of NL questions and database schema, indicating that existing ICL-based techniques fail to augment LLMs in mapping NL to concrete database objects. The impact of LLM capability is more obvious on the semantic errors. Replacing GPT-3.5 with the more powerful GPT-4o model, the studied techniques have 48.7-80.4\% less semantic errors.

Around 40\% semantic errors are \emph{projection errors}. 
The generated queries frequently use a wrong numerator or denominator in a division operation, especially for ratio calculations. Similar problems are also observed in other arithmetic-based logic. These demonstrate the limitations of existing LLMs in mathematical reasoning within the context of relational structures and query formulations.

\emph{Incorrect table selection} errors account for 22.4\% and 35.0\% of semantic errors in \bird and \spider, respectively. In complex databases, LLMs are likely to select wrong tables from a set of correlated ones. For example, if two tables share similar column names, it is hard for LLMs to distinguish them from the semantic space.
The high frequency of \emph{incorrect table selection} highlights the gap between natural languages and programming languages.

\begin{tcolorbox}[width=\linewidth, boxrule=0pt, sharp corners=all, left=2pt, right=2pt, top=0pt, bottom=0pt, colback=gray!20]
\textbf{Finding 4}: Semantic errors account for 30.9\% errors. They are more common than intent-independent errors when solving harder tasks (\ie, \bird). They are caused by the misinterpretation of natural languages and misunderstanding of the database schema.
\end{tcolorbox}

\subsubsection{Others.}\hfill

As introduced in Section~\ref{sec:others_error}, more than a third of errors cannot be classified by our taxonomy to provide insights for detection and repairing solution. It indicates that the ICL-based techniques generate completely wrong SQL queries. We believe that these errors should be resolved by technique re-design or LLM enhancement, instead of run-time repairing.
\edit{Meanwhile, There are also 120 errors clustered into \emph{miscellaneous error}.}

\subsection{Discussion}

\subsubsection{Quality of Benchmarks.} \hfill
\label{sec:error_bench_quality}

The significant amount of \emph{not-an-error} problems highlights the benchmark quality problems. In all the 12340 generated queries, 1905 have unreliable correctness judgments. 
Specifically, 52.1\% of them are caused by the incorrect ground-truth (\ie, gold errors), indicating prevalent human annotation mistakes. 
The imprecise database schema description and ambiguous NL questions also cause \emph{not-an-error} problems. 
We observe severe foreign key violations in the \code{flight\_2} database of \spider, and thus omit this database in the evaluation of Section~\ref{sec:method_effectiveness}\&\ref{sec:tool} to improve reliability. 

While our study has identified and reported many \emph{not-an-error} problems 
to the benchmark developers, there might still remain missing ones. The generated query may make the same mistakes as the ground-truth, and thus be regarded as correct by the \emph{execution match} metrics, which we do not focus on. Therefore, we call for a high-quality text-to-SQL benchmark.

\subsubsection{Differences of Benchmarks.} \hfill

We observe notable differences in error distributions between benchmarks. 
The main reason is that \bird and \spider have different distributions of problem types, directly affecting task difficulty.
Particularly, different problem types may require different SQL keywords, leading to varied errors. For example, \spider contains many set operations, while \bird does not.
In addition, \spider does not provide evidence and thus has less misleading information, reducing the likelihood of errors like \emph{function hallucination}, and \emph{alias not used}. 
These findings further prove the importance of understanding benchmark characteristics when interpreting evaluation results.

\subsubsection{Capability of LLMs.} \hfill

Switching from GPT-3.5 to GPT-4o, the ICL-based techniques reduce 42.7\% errors. Specifically, MAC-SQL reduced more than half errors in \bird, and three quarters in \spider. 
% 1-(448+108+84)/(786+188+143)=0.427
It shows that LLM capability greatly affects technique performance, and the bottleneck has not been reached yet.

Meanwhile, for some error types, switching to a more powerful LLM either does not reduce errors or even increases them. It is particularly visible in \emph{using = instead of IN}, \emph{ascending sort with NULL}, and other error types with strong logic features.

\subsubsection{Symptoms of Errors.} \label{sec:consequences_error} \hfill 

Syntax and schema errors prevent the SQL query from executing, making them observable. Logic and convention errors typically exhibit clear anti-patterns in the query. In contrast, semantic errors are harder to detect and require NL interpretation. \edit{For semantic errors, we rely on error propagation. Semantic faults (\eg, \textit{Improper Condition}) often inadvertently filter out all valid data. Consequently, these errors converge on the same ``suspicious symptoms'' of empty result sets and NULL results, which serve as effective proxies for detecting semantic errors.}

%% file: tables/chapter_3_detailed_error_statistics.tex
% Please add the following required packages to your document preamble:
% \usepackage{multirow}
% \usepackage[table,xcdraw]{xcolor}
% Beamer presentation requires \usepackage{colortbl} instead of \usepackage[table,xcdraw]{xcolor}
\begin{table*}[]
\centering
\footnotesize
\caption{
    Number of errors of ICL-based text-to-SQL techniques. 
    % \shen{Script to fetch statistics can used in reproducibility check now.}
    % \wan{Total numbers of all errors should be more eye-catching}
}
\vspace{-0.1in}
\label{table: error result}
\setlength{\tabcolsep}{2pt}
\resizebox{\linewidth}{!}{%
\begin{threeparttable}[b]
\begin{tabular}{|l|cccc|cccccc||cccccc|}
\hline
\multicolumn{1}{|c|}{}                   & \multicolumn{4}{c|}{BIRD}                                                                                                                                                                        & \multicolumn{6}{c||}{SPIDER}                                                                                                                                                                                                                                                                                              & \multicolumn{6}{c|}{BIRD, Repairing Task}                                                                                                                                                                                                                                                                                                \\ \cline{2-17} 
\multicolumn{1}{|c|}{}                   & \multicolumn{3}{c|}{GPT-3.5-Turbo}                                                                                                                                         & GPT-4o       & \multicolumn{4}{c|}{GPT-3.5-Turbo}                                                                                                                                                                                                     & \multicolumn{2}{c||}{GPT-4o}                                              & \multicolumn{6}{c|}{GPT-3.5-Turbo}                                                                                                                                                                                                                                                                                \\ \cline{2-17} 
\multicolumn{1}{|c|}{\multirow{-3}{*}{}} & \multicolumn{1}{c|}{MAC-SQL}                              & \multicolumn{1}{c|}{DIN-SQL}                              & \multicolumn{1}{c|}{CHESS}                                & MAC-SQL      & \multicolumn{1}{c|}{MAC-SQL}                              & \multicolumn{1}{c|}{DIN-SQL}                              & \multicolumn{1}{c|}{CHESS}                                & \multicolumn{1}{c|}{DEA-SQL}                              & \multicolumn{1}{c|}{MAC-SQL}                              & DEA-SQL      & \multicolumn{1}{c|}{Before}                               & \multicolumn{1}{c|}{Rule-Exe}                             & \multicolumn{1}{c|}{LLM-Plain}                            & \multicolumn{1}{c|}{LLM-Exe}                              & \multicolumn{1}{c|}{LLM-Value}                            & LLM-Extr     \\ \hline
Function Hallucination                   & \multicolumn{1}{c|}{8}                                    & \multicolumn{1}{c|}{2}                                    & \multicolumn{1}{c|}{17}                                   & 0            & \multicolumn{1}{c|}{0}                                    & \multicolumn{1}{c|}{0}                                    & \multicolumn{1}{c|}{0}                                    & \multicolumn{1}{c|}{0}                                    & \multicolumn{1}{c|}{0}                                    & 0            & \multicolumn{1}{c|}{8}                                    & \multicolumn{1}{c|}{0}                                    & \multicolumn{1}{c|}{4}                                    & \multicolumn{1}{c|}{1}                                    & \multicolumn{1}{c|}{1}                                    & 4            \\ \hline
Missing Quote                            & \multicolumn{1}{c|}{1}                                    & \multicolumn{1}{c|}{8}                                    & \multicolumn{1}{c|}{1}                                    & 0            & \multicolumn{1}{c|}{0}                                    & \multicolumn{1}{c|}{0}                                    & \multicolumn{1}{c|}{0}                                    & \multicolumn{1}{c|}{0}                                    & \multicolumn{1}{c|}{0}                                    & 0            & \multicolumn{1}{c|}{1}                                    & \multicolumn{1}{c|}{0}                                    & \multicolumn{1}{c|}{0}                                    & \multicolumn{1}{c|}{0}                                    & \multicolumn{1}{c|}{0}                                    & 0            \\ \hline
Other Syntax Violations                  & \multicolumn{1}{c|}{8}                                    & \multicolumn{1}{c|}{10}                                   & \multicolumn{1}{c|}{5}                                    & 2            & \multicolumn{1}{c|}{1}                                    & \multicolumn{1}{c|}{0}                                    & \multicolumn{1}{c|}{7}                                    & \multicolumn{1}{c|}{2}                                    & \multicolumn{1}{c|}{0}                                    & 1            & \multicolumn{1}{c|}{8}                                    & \multicolumn{1}{c|}{5}                                    & \multicolumn{1}{c|}{9}                                    & \multicolumn{1}{c|}{8}                                    & \multicolumn{1}{c|}{7}                                    & 10           \\ \hline
\rowcolor[HTML]{D9D9D9} 
\textbf{SUM: Syntax Error}               & \multicolumn{1}{c|}{\cellcolor[HTML]{D9D9D9}\textbf{17}}  & \multicolumn{1}{c|}{\cellcolor[HTML]{D9D9D9}\textbf{20}}  & \multicolumn{1}{c|}{\cellcolor[HTML]{D9D9D9}\textbf{23}}  & \textbf{2}   & \multicolumn{1}{c|}{\cellcolor[HTML]{D9D9D9}\textbf{1}}   & \multicolumn{1}{c|}{\cellcolor[HTML]{D9D9D9}\textbf{0}}   & \multicolumn{1}{c|}{\cellcolor[HTML]{D9D9D9}\textbf{7}}   & \multicolumn{1}{c|}{\cellcolor[HTML]{D9D9D9}\textbf{2}}   & \multicolumn{1}{c|}{\cellcolor[HTML]{D9D9D9}\textbf{0}}   & \textbf{1}   & \multicolumn{1}{c|}{\cellcolor[HTML]{D9D9D9}\textbf{17}}  & \multicolumn{1}{c|}{\cellcolor[HTML]{D9D9D9}\textbf{5}}   & \multicolumn{1}{c|}{\cellcolor[HTML]{D9D9D9}\textbf{13}}  & \multicolumn{1}{c|}{\cellcolor[HTML]{D9D9D9}\textbf{9}}   & \multicolumn{1}{c|}{\cellcolor[HTML]{D9D9D9}\textbf{8}}   & \textbf{14}  \\ \hline \hline
Table-Column Mismatch                    & \multicolumn{1}{c|}{70}                                   & \multicolumn{1}{c|}{39}                                   & \multicolumn{1}{c|}{22}                                   & 13           & \multicolumn{1}{c|}{16}                                   & \multicolumn{1}{c|}{8}                                    & \multicolumn{1}{c|}{3}                                    & \multicolumn{1}{c|}{5}                                    & \multicolumn{1}{c|}{0}                                    & 0            & \multicolumn{1}{c|}{70}                                   & \multicolumn{1}{c|}{22}                                   & \multicolumn{1}{c|}{43}                                   & \multicolumn{1}{c|}{34}                                   & \multicolumn{1}{c|}{33}                                   & 42           \\ \hline
Unused Alias                             & \multicolumn{1}{c|}{0}                                    & \multicolumn{1}{c|}{0}                                    & \multicolumn{1}{c|}{14}                                   & 0            & \multicolumn{1}{c|}{0}                                    & \multicolumn{1}{c|}{0}                                    & \multicolumn{1}{c|}{2}                                    & \multicolumn{1}{c|}{0}                                    & \multicolumn{1}{c|}{0}                                    & 0            & \multicolumn{1}{c|}{0}                                    & \multicolumn{1}{c|}{0}                                    & \multicolumn{1}{c|}{0}                                    & \multicolumn{1}{c|}{0}                                    & \multicolumn{1}{c|}{0}                                    & 0            \\ \hline
Schema Hallucination                     & \multicolumn{1}{c|}{8}                                    & \multicolumn{1}{c|}{7}                                    & \multicolumn{1}{c|}{16}                                   & 2            & \multicolumn{1}{c|}{4}                                    & \multicolumn{1}{c|}{2}                                    & \multicolumn{1}{c|}{24}                                   & \multicolumn{1}{c|}{4}                                    & \multicolumn{1}{c|}{0}                                    & 0            & \multicolumn{1}{c|}{8}                                    & \multicolumn{1}{c|}{2}                                    & \multicolumn{1}{c|}{4}                                    & \multicolumn{1}{c|}{3}                                    & \multicolumn{1}{c|}{3}                                    & 3            \\ \hline
Ambiguous Reference                      & \multicolumn{1}{c|}{1}                                    & \multicolumn{1}{c|}{2}                                    & \multicolumn{1}{c|}{3}                                    & 0            & \multicolumn{1}{c|}{3}                                    & \multicolumn{1}{c|}{33}                                   & \multicolumn{1}{c|}{3}                                    & \multicolumn{1}{c|}{12}                                   & \multicolumn{1}{c|}{0}                                    & 0            & \multicolumn{1}{c|}{1}                                    & \multicolumn{1}{c|}{0}                                    & \multicolumn{1}{c|}{0}                                    & \multicolumn{1}{c|}{0}                                    & \multicolumn{1}{c|}{0}                                    & 0            \\ \hline
\rowcolor[HTML]{D9D9D9} 
\textbf{SUM: Schema Error}               & \multicolumn{1}{c|}{\cellcolor[HTML]{D9D9D9}\textbf{79}}  & \multicolumn{1}{c|}{\cellcolor[HTML]{D9D9D9}\textbf{48}}  & \multicolumn{1}{c|}{\cellcolor[HTML]{D9D9D9}\textbf{55}}  & \textbf{15}  & \multicolumn{1}{c|}{\cellcolor[HTML]{D9D9D9}\textbf{23}}  & \multicolumn{1}{c|}{\cellcolor[HTML]{D9D9D9}\textbf{43}}  & \multicolumn{1}{c|}{\cellcolor[HTML]{D9D9D9}\textbf{32}}  & \multicolumn{1}{c|}{\cellcolor[HTML]{D9D9D9}\textbf{21}}  & \multicolumn{1}{c|}{\cellcolor[HTML]{D9D9D9}\textbf{0}}   & \textbf{0}   & \multicolumn{1}{c|}{\cellcolor[HTML]{D9D9D9}\textbf{79}}  & \multicolumn{1}{c|}{\cellcolor[HTML]{D9D9D9}\textbf{24}}  & \multicolumn{1}{c|}{\cellcolor[HTML]{D9D9D9}\textbf{47}}  & \multicolumn{1}{c|}{\cellcolor[HTML]{D9D9D9}\textbf{37}}  & \multicolumn{1}{c|}{\cellcolor[HTML]{D9D9D9}\textbf{36}}  & \textbf{45}  \\ \hline \hline
Implicit Type Conversion                 & \multicolumn{1}{c|}{6}                                    & \multicolumn{1}{c|}{7}                                    & \multicolumn{1}{c|}{8}                                    & 0            & \multicolumn{1}{c|}{0}                                    & \multicolumn{1}{c|}{0}                                    & \multicolumn{1}{c|}{0}                                    & \multicolumn{1}{c|}{0}                                    & \multicolumn{1}{c|}{0}                                    & 0            & \multicolumn{1}{c|}{6}                                    & \multicolumn{1}{c|}{6}                                    & \multicolumn{1}{c|}{5}                                    & \multicolumn{1}{c|}{6}                                    & \multicolumn{1}{c|}{6}                                    & 5            \\ \hline
Using = instead of IN                    & \multicolumn{1}{c|}{15}                                   & \multicolumn{1}{c|}{1}                                    & \multicolumn{1}{c|}{0}                                    & 22           & \multicolumn{1}{c|}{1}                                    & \multicolumn{1}{c|}{1}                                    & \multicolumn{1}{c|}{1}                                    & \multicolumn{1}{c|}{0}                                    & \multicolumn{1}{c|}{0}                                    & 0            & \multicolumn{1}{c|}{15}                                   & \multicolumn{1}{c|}{9}                                    & \multicolumn{1}{c|}{0}                                    & \multicolumn{1}{c|}{0}                                    & \multicolumn{1}{c|}{0}                                    & 0            \\ \hline
Ascending Sort with NULL                 & \multicolumn{1}{c|}{2}                                    & \multicolumn{1}{c|}{3}                                    & \multicolumn{1}{c|}{3}                                    & 5            & \multicolumn{1}{c|}{0}                                    & \multicolumn{1}{c|}{0}                                    & \multicolumn{1}{c|}{0}                                    & \multicolumn{1}{c|}{0}                                    & \multicolumn{1}{c|}{0}                                    & 0            & \multicolumn{1}{c|}{2}                                    & \multicolumn{1}{c|}{2}                                    & \multicolumn{1}{c|}{2}                                    & \multicolumn{1}{c|}{2}                                    & \multicolumn{1}{c|}{2}                                    & 2            \\ \hline
\rowcolor[HTML]{D9D9D9} 
\textbf{SUM: Logic Error}                & \multicolumn{1}{c|}{\cellcolor[HTML]{D9D9D9}\textbf{23}}  & \multicolumn{1}{c|}{\cellcolor[HTML]{D9D9D9}\textbf{11}}  & \multicolumn{1}{c|}{\cellcolor[HTML]{D9D9D9}\textbf{11}}  & \textbf{27}  & \multicolumn{1}{c|}{\cellcolor[HTML]{D9D9D9}\textbf{1}}   & \multicolumn{1}{c|}{\cellcolor[HTML]{D9D9D9}\textbf{1}}   & \multicolumn{1}{c|}{\cellcolor[HTML]{D9D9D9}\textbf{1}}   & \multicolumn{1}{c|}{\cellcolor[HTML]{D9D9D9}\textbf{0}}   & \multicolumn{1}{c|}{\cellcolor[HTML]{D9D9D9}\textbf{0}}   & \textbf{0}   & \multicolumn{1}{c|}{\cellcolor[HTML]{D9D9D9}\textbf{23}}  & \multicolumn{1}{c|}{\cellcolor[HTML]{D9D9D9}\textbf{17}}  & \multicolumn{1}{c|}{\cellcolor[HTML]{D9D9D9}\textbf{7}}   & \multicolumn{1}{c|}{\cellcolor[HTML]{D9D9D9}\textbf{8}}   & \multicolumn{1}{c|}{\cellcolor[HTML]{D9D9D9}\textbf{8}}   & \textbf{7}   \\ \hline \hline
Violating Value Specification            & \multicolumn{1}{c|}{59}                                   & \multicolumn{1}{c|}{60}                                   & \multicolumn{1}{c|}{88}                                   & 26           & \multicolumn{1}{c|}{10}                                   & \multicolumn{1}{c|}{40}                                   & \multicolumn{1}{c|}{30}                                   & \multicolumn{1}{c|}{11}                                   & \multicolumn{1}{c|}{8}                                    & 7            & \multicolumn{1}{c|}{59}                                   & \multicolumn{1}{c|}{45}                                   & \multicolumn{1}{c|}{40}                                   & \multicolumn{1}{c|}{43}                                   & \multicolumn{1}{c|}{33}                                   & 40           \\ \hline
Aggregation/Comparison Misuse            & \multicolumn{1}{c|}{8}                                    & \multicolumn{1}{c|}{5}                                    & \multicolumn{1}{c|}{7}                                    & 4            & \multicolumn{1}{c|}{0}                                    & \multicolumn{1}{c|}{0}                                    & \multicolumn{1}{c|}{0}                                    & \multicolumn{1}{c|}{0}                                    & \multicolumn{1}{c|}{0}                                    & 0            & \multicolumn{1}{c|}{8}                                    & \multicolumn{1}{c|}{4}                                    & \multicolumn{1}{c|}{3}                                    & \multicolumn{1}{c|}{3}                                    & \multicolumn{1}{c|}{3}                                    & 1            \\ \hline
Comparing Unrelated Columns              & \multicolumn{1}{c|}{6}                                    & \multicolumn{1}{c|}{27}                                   & \multicolumn{1}{c|}{58}                                   & 3            & \multicolumn{1}{c|}{6}                                    & \multicolumn{1}{c|}{14}                                   & \multicolumn{1}{c|}{3}                                    & \multicolumn{1}{c|}{2}                                    & \multicolumn{1}{c|}{3}                                    & 0            & \multicolumn{1}{c|}{6}                                    & \multicolumn{1}{c|}{2}                                    & \multicolumn{1}{c|}{2}                                    & \multicolumn{1}{c|}{1}                                    & \multicolumn{1}{c|}{2}                                    & 2            \\ \hline
\rowcolor[HTML]{D9D9D9} 
\textbf{SUM: Convention Error}           & \multicolumn{1}{c|}{\cellcolor[HTML]{D9D9D9}\textbf{73}}  & \multicolumn{1}{c|}{\cellcolor[HTML]{D9D9D9}\textbf{92}}  & \multicolumn{1}{c|}{\cellcolor[HTML]{D9D9D9}\textbf{153}} & \textbf{33}  & \multicolumn{1}{c|}{\cellcolor[HTML]{D9D9D9}\textbf{16}}  & \multicolumn{1}{c|}{\cellcolor[HTML]{D9D9D9}\textbf{54}}  & \multicolumn{1}{c|}{\cellcolor[HTML]{D9D9D9}\textbf{33}}  & \multicolumn{1}{c|}{\cellcolor[HTML]{D9D9D9}\textbf{13}}  & \multicolumn{1}{c|}{\cellcolor[HTML]{D9D9D9}\textbf{11}}  & \textbf{7}   & \multicolumn{1}{c|}{\cellcolor[HTML]{D9D9D9}\textbf{73}}  & \multicolumn{1}{c|}{\cellcolor[HTML]{D9D9D9}\textbf{51}}  & \multicolumn{1}{c|}{\cellcolor[HTML]{D9D9D9}\textbf{45}}  & \multicolumn{1}{c|}{\cellcolor[HTML]{D9D9D9}\textbf{47}}  & \multicolumn{1}{c|}{\cellcolor[HTML]{D9D9D9}\textbf{38}}  & \textbf{43}  \\ \hline \hline
Wrong COUNT Object                       & \multicolumn{1}{c|}{10}                                   & \multicolumn{1}{c|}{10}                                   & \multicolumn{1}{c|}{16}                                   & 7            & \multicolumn{1}{c|}{0}                                    & \multicolumn{1}{c|}{2}                                    & \multicolumn{1}{c|}{0}                                    & \multicolumn{1}{c|}{1}                                    & \multicolumn{1}{c|}{0}                                    & 0            & \multicolumn{1}{c|}{10}                                   & \multicolumn{1}{c|}{9}                                    & \multicolumn{1}{c|}{9}                                    & \multicolumn{1}{c|}{7}                                    & \multicolumn{1}{c|}{8}                                    & 9            \\ \hline
Subquery Scope Inconsistency             & \multicolumn{1}{c|}{5}                                    & \multicolumn{1}{c|}{1}                                    & \multicolumn{1}{c|}{6}                                    & 0            & \multicolumn{1}{c|}{0}                                    & \multicolumn{1}{c|}{0}                                    & \multicolumn{1}{c|}{0}                                    & \multicolumn{1}{c|}{0}                                    & \multicolumn{1}{c|}{0}                                    & 0            & \multicolumn{1}{c|}{5}                                    & \multicolumn{1}{c|}{3}                                    & \multicolumn{1}{c|}{3}                                    & \multicolumn{1}{c|}{3}                                    & \multicolumn{1}{c|}{3}                                    & 4            \\ \hline
Missing DISTINCT Keywords                & \multicolumn{1}{c|}{4}                                    & \multicolumn{1}{c|}{6}                                    & \multicolumn{1}{c|}{13}                                   & 5            & \multicolumn{1}{c|}{0}                                    & \multicolumn{1}{c|}{0}                                    & \multicolumn{1}{c|}{0}                                    & \multicolumn{1}{c|}{0}                                    & \multicolumn{1}{c|}{0}                                    & 0            & \multicolumn{1}{c|}{4}                                    & \multicolumn{1}{c|}{4}                                    & \multicolumn{1}{c|}{7}                                    & \multicolumn{1}{c|}{7}                                    & \multicolumn{1}{c|}{6}                                    & 7            \\ \hline
Comparing Wrong Columns                  & \multicolumn{1}{c|}{3}                                    & \multicolumn{1}{c|}{0}                                    & \multicolumn{1}{c|}{2}                                    & 1            & \multicolumn{1}{c|}{0}                                    & \multicolumn{1}{c|}{0}                                    & \multicolumn{1}{c|}{0}                                    & \multicolumn{1}{c|}{0}                                    & \multicolumn{1}{c|}{0}                                    & 0            & \multicolumn{1}{c|}{3}                                    & \multicolumn{1}{c|}{3}                                    & \multicolumn{1}{c|}{1}                                    & \multicolumn{1}{c|}{2}                                    & \multicolumn{1}{c|}{2}                                    & 1            \\ \hline
Projection Error                         & \multicolumn{1}{c|}{115}                                  & \multicolumn{1}{c|}{84}                                   & \multicolumn{1}{c|}{100}                                  & 84           & \multicolumn{1}{c|}{10}                                   & \multicolumn{1}{c|}{2}                                    & \multicolumn{1}{c|}{21}                                   & \multicolumn{1}{c|}{3}                                    & \multicolumn{1}{c|}{1}                                    & 1            & \multicolumn{1}{c|}{115}                                  & \multicolumn{1}{c|}{111}                                  & \multicolumn{1}{c|}{83}                                   & \multicolumn{1}{c|}{88}                                   & \multicolumn{1}{c|}{85}                                   & 83           \\ \hline
ORDER BY Errors                          & \multicolumn{1}{c|}{6}                                    & \multicolumn{1}{c|}{4}                                    & \multicolumn{1}{c|}{2}                                    & 1            & \multicolumn{1}{c|}{0}                                    & \multicolumn{1}{c|}{1}                                    & \multicolumn{1}{c|}{7}                                    & \multicolumn{1}{c|}{0}                                    & \multicolumn{1}{c|}{0}                                    & 0            & \multicolumn{1}{c|}{6}                                    & \multicolumn{1}{c|}{8}                                    & \multicolumn{1}{c|}{5}                                    & \multicolumn{1}{c|}{7}                                    & \multicolumn{1}{c|}{7}                                    & 9            \\ \hline
Improper Condition                       & \multicolumn{1}{c|}{64}                                   & \multicolumn{1}{c|}{31}                                   & \multicolumn{1}{c|}{68}                                   & 42           & \multicolumn{1}{c|}{9}                                    & \multicolumn{1}{c|}{3}                                    & \multicolumn{1}{c|}{15}                                   & \multicolumn{1}{c|}{0}                                    & \multicolumn{1}{c|}{2}                                    & 2            & \multicolumn{1}{c|}{64}                                   & \multicolumn{1}{c|}{55}                                   & \multicolumn{1}{c|}{39}                                   & \multicolumn{1}{c|}{45}                                   & \multicolumn{1}{c|}{49}                                   & 39           \\ \hline
Incorrect Table Selection                & \multicolumn{1}{c|}{87}                                   & \multicolumn{1}{c|}{43}                                   & \multicolumn{1}{c|}{72}                                   & 10           & \multicolumn{1}{c|}{19}                                   & \multicolumn{1}{c|}{17}                                   & \multicolumn{1}{c|}{17}                                   & \multicolumn{1}{c|}{6}                                    & \multicolumn{1}{c|}{5}                                    & 0            & \multicolumn{1}{c|}{87}                                   & \multicolumn{1}{c|}{71}                                   & \multicolumn{1}{c|}{67}                                   & \multicolumn{1}{c|}{65}                                   & \multicolumn{1}{c|}{64}                                   & 67           \\ \hline
Unaligned Aggregation Structure          & \multicolumn{1}{c|}{10}                                   & \multicolumn{1}{c|}{14}                                   & \multicolumn{1}{c|}{14}                                   & 6            & \multicolumn{1}{c|}{8}                                    & \multicolumn{1}{c|}{16}                                   & \multicolumn{1}{c|}{5}                                    & \multicolumn{1}{c|}{7}                                    & \multicolumn{1}{c|}{1}                                    & 2            & \multicolumn{1}{c|}{10}                                   & \multicolumn{1}{c|}{11}                                   & \multicolumn{1}{c|}{7}                                    & \multicolumn{1}{c|}{8}                                    & \multicolumn{1}{c|}{8}                                    & 9            \\ \hline
\rowcolor[HTML]{D9D9D9} 
\textbf{SUM: Semantic Error}             & \multicolumn{1}{c|}{\cellcolor[HTML]{D9D9D9}\textbf{304}} & \multicolumn{1}{c|}{\cellcolor[HTML]{D9D9D9}\textbf{193}} & \multicolumn{1}{c|}{\cellcolor[HTML]{D9D9D9}\textbf{293}} & \textbf{156} & \multicolumn{1}{c|}{\cellcolor[HTML]{D9D9D9}\textbf{46}}  & \multicolumn{1}{c|}{\cellcolor[HTML]{D9D9D9}\textbf{41}}  & \multicolumn{1}{c|}{\cellcolor[HTML]{D9D9D9}\textbf{65}}  & \multicolumn{1}{c|}{\cellcolor[HTML]{D9D9D9}\textbf{17}}  & \multicolumn{1}{c|}{\cellcolor[HTML]{D9D9D9}\textbf{9}}   & \textbf{5}   & \multicolumn{1}{c|}{\cellcolor[HTML]{D9D9D9}\textbf{304}} & \multicolumn{1}{c|}{\cellcolor[HTML]{D9D9D9}\textbf{275}} & \multicolumn{1}{c|}{\cellcolor[HTML]{D9D9D9}\textbf{221}} & \multicolumn{1}{c|}{\cellcolor[HTML]{D9D9D9}\textbf{232}} & \multicolumn{1}{c|}{\cellcolor[HTML]{D9D9D9}\textbf{232}} & \textbf{228} \\ \hline \hline
Unclassifiable                           & \multicolumn{1}{c|}{274}                                  & \multicolumn{1}{c|}{240}                                  & \multicolumn{1}{c|}{211}                                  & 195          & \multicolumn{1}{c|}{97}                                   & \multicolumn{1}{c|}{98}                                   & \multicolumn{1}{c|}{121}                                  & \multicolumn{1}{c|}{77}                                   & \multicolumn{1}{c|}{83}                                   & 61           & \multicolumn{1}{c|}{274}                                  & \multicolumn{1}{c|}{270}                                  & \multicolumn{1}{c|}{296}                                  & \multicolumn{1}{c|}{273}                                  & \multicolumn{1}{c|}{275}                                  & 289          \\ \hline
Miscellaneous Error                      & \multicolumn{1}{c|}{16}                                   & \multicolumn{1}{c|}{17}                                   & \multicolumn{1}{c|}{19}                                   & 20           & \multicolumn{1}{c|}{4}                                    & \multicolumn{1}{c|}{3}                                    & \multicolumn{1}{c|}{13}                                   & \multicolumn{1}{c|}{13}                                   & \multicolumn{1}{c|}{5}                                    & 10           & \multicolumn{1}{c|}{16}                                   & \multicolumn{1}{c|}{15}                                   & \multicolumn{1}{c|}{12}                                   & \multicolumn{1}{c|}{20}                                   & \multicolumn{1}{c|}{16}                                   & 28           \\ \hline
\rowcolor[HTML]{D9D9D9} 
\textbf{SUM: Others}                     & \multicolumn{1}{c|}{\cellcolor[HTML]{D9D9D9}\textbf{290}} & \multicolumn{1}{c|}{\cellcolor[HTML]{D9D9D9}\textbf{257}} & \multicolumn{1}{c|}{\cellcolor[HTML]{D9D9D9}\textbf{230}} & \textbf{215} & \multicolumn{1}{c|}{\cellcolor[HTML]{D9D9D9}\textbf{101}} & \multicolumn{1}{c|}{\cellcolor[HTML]{D9D9D9}\textbf{101}} & \multicolumn{1}{c|}{\cellcolor[HTML]{D9D9D9}\textbf{134}} & \multicolumn{1}{c|}{\cellcolor[HTML]{D9D9D9}\textbf{90}}  & \multicolumn{1}{c|}{\cellcolor[HTML]{D9D9D9}\textbf{88}}  & \textbf{71}  & \multicolumn{1}{c|}{\cellcolor[HTML]{D9D9D9}\textbf{290}} & \multicolumn{1}{c|}{\cellcolor[HTML]{D9D9D9}\textbf{285}} & \multicolumn{1}{c|}{\cellcolor[HTML]{D9D9D9}\textbf{308}} & \multicolumn{1}{c|}{\cellcolor[HTML]{D9D9D9}\textbf{293}} & \multicolumn{1}{c|}{\cellcolor[HTML]{D9D9D9}\textbf{291}} & \textbf{317} \\ \hline \hline
\rowcolor[HTML]{A6A6A6} 
\textbf{Total number of all errors}      & \multicolumn{1}{c|}{\cellcolor[HTML]{A6A6A6}\textbf{786}} & \multicolumn{1}{c|}{\cellcolor[HTML]{A6A6A6}\textbf{621}} & \multicolumn{1}{c|}{\cellcolor[HTML]{A6A6A6}\textbf{765}} & \textbf{448} & \multicolumn{1}{c|}{\cellcolor[HTML]{A6A6A6}\textbf{188}} & \multicolumn{1}{c|}{\cellcolor[HTML]{A6A6A6}\textbf{240}} & \multicolumn{1}{c|}{\cellcolor[HTML]{A6A6A6}\textbf{272}} & \multicolumn{1}{c|}{\cellcolor[HTML]{A6A6A6}\textbf{143}} & \multicolumn{1}{c|}{\cellcolor[HTML]{A6A6A6}\textbf{108}} & \textbf{84}  & \multicolumn{1}{c|}{\cellcolor[HTML]{A6A6A6}\textbf{786}} & \multicolumn{1}{c|}{\cellcolor[HTML]{A6A6A6}\textbf{657}} & \multicolumn{1}{c|}{\cellcolor[HTML]{A6A6A6}\textbf{641}} & \multicolumn{1}{c|}{\cellcolor[HTML]{A6A6A6}\textbf{626}} & \multicolumn{1}{c|}{\cellcolor[HTML]{A6A6A6}\textbf{613}} & \textbf{654} \\ \hline \hline \hline
Output Ambiguity                            & \multicolumn{1}{c|}{79}                                   & \multicolumn{1}{c|}{123}                                   & \multicolumn{1}{c|}{123}                                  & 112           & \multicolumn{1}{c|}{96}                                   & \multicolumn{1}{c|}{81}                                   & \multicolumn{1}{c|}{70}                                   & \multicolumn{1}{c|}{84}                                   & \multicolumn{1}{c|}{98}                                   & 78           & \multicolumn{1}{c|}{79}                                   & \multicolumn{1}{c|}{77}                                   & \multicolumn{1}{c|}{68}                                   & \multicolumn{1}{c|}{73}                                   & \multicolumn{1}{c|}{73}                                   & 70           \\ \hline
Gold Error                               & \multicolumn{1}{c|}{176}                                  & \multicolumn{1}{c|}{158}                                  & \multicolumn{1}{c|}{159}                                  & 179          & \multicolumn{1}{c|}{50}                                   & \multicolumn{1}{c|}{56}                                   & \multicolumn{1}{c|}{56}                                   & \multicolumn{1}{c|}{53}                                   & \multicolumn{1}{c|}{54}                                   & 52           & \multicolumn{1}{c|}{176}                                  & \multicolumn{1}{c|}{173}                                  & \multicolumn{1}{c|}{169}                                  & \multicolumn{1}{c|}{163}                                  & \multicolumn{1}{c|}{160}                                  & 168          \\ \hline
Violation of Foreign Key Integrity       & \multicolumn{1}{c|}{11}                                   & \multicolumn{1}{c|}{9}                                    & \multicolumn{1}{c|}{6}                                    & 5            & \multicolumn{1}{c|}{0}                                    & \multicolumn{1}{c|}{2}                                    & \multicolumn{1}{c|}{0}                                    & \multicolumn{1}{c|}{0}                                    & \multicolumn{1}{c|}{0}                                    & 0            & \multicolumn{1}{c|}{11}                                   & \multicolumn{1}{c|}{9}                                    & \multicolumn{1}{c|}{7}                                    & \multicolumn{1}{c|}{8}                                    & \multicolumn{1}{c|}{9}                                    & 7            \\ \hline
\rowcolor[HTML]{D9D9D9} 
\textbf{SUM: Not an Error}               & \multicolumn{1}{c|}{\cellcolor[HTML]{D9D9D9}\textbf{266}} & \multicolumn{1}{c|}{\cellcolor[HTML]{D9D9D9}\textbf{290}} & \multicolumn{1}{c|}{\cellcolor[HTML]{D9D9D9}\textbf{288}} & \textbf{296} & \multicolumn{1}{c|}{\cellcolor[HTML]{D9D9D9}\textbf{146}} & \multicolumn{1}{c|}{\cellcolor[HTML]{D9D9D9}\textbf{139}} & \multicolumn{1}{c|}{\cellcolor[HTML]{D9D9D9}\textbf{126}} & \multicolumn{1}{c|}{\cellcolor[HTML]{D9D9D9}\textbf{137}} & \multicolumn{1}{c|}{\cellcolor[HTML]{D9D9D9}\textbf{152}} & \textbf{130} & \multicolumn{1}{c|}{\cellcolor[HTML]{D9D9D9}\textbf{266}} & \multicolumn{1}{c|}{\cellcolor[HTML]{D9D9D9}\textbf{259}} & \multicolumn{1}{c|}{\cellcolor[HTML]{D9D9D9}\textbf{244}} & \multicolumn{1}{c|}{\cellcolor[HTML]{D9D9D9}\textbf{244}} & \multicolumn{1}{c|}{\cellcolor[HTML]{D9D9D9}\textbf{242}} & \textbf{245} \\ \hline
\end{tabular}
\begin{tablenotes}
\item [1] Left part shows the error statistics of ICL-based techniques before repairing. Right part shows error statistics after repairing.
\end{tablenotes}
\end{threeparttable}%
}
\vspace{-0.2in}
\end{table*}

%% file: sections/4-Effectiveness.tex
\label{sec:method_effectiveness}

\subsection{Methodology}

\paragraph{Repairing Framework.} We develop a fixing framework to evaluate existing repairing methods. It is built upon MAC-SQL due to its straightforward pipeline. The prompt template consists of five components: (1) schema representation (with clearer foreign key information), (2) question and evidence, (3) general instructions (using CHESS's design for broader error coverage), (4) SQL query to be repaired, and (5) supplementary information if needed.

\paragraph{Schemes.} We summarize five basic repairing methods from the repairing module of the four techniques evaluated in Section~\ref{sec:errors}, as shown in Table~\ref{table:Effectiveness Setup}. 
\begin{itemize} [leftmargin=*]
    \item \emph{Rule-Exe}: It applies rule-based error detection according to the SQL execution results. It asks LLMs to regenerate the query if the original one can not be executed, returns empty result, or contains \code{NULL} values in results.
    \item \emph{LLM-Plain}: For each generated SQL query, the LLM is asked to regenerate without additional information.
    \item \emph{LLM-Exe}: For each query, the LLM regenerates SQL query according to the execution results.
    \item \emph{LLM-Value}: It extends \emph{LLM-Exe} by providing value specification.
    \item \emph{LLM-Extr}: It extends LLM-Plain by appending a second round fixing with instructions and few-shot repairing examples for extremum-related questions {(see F3: Output Ambiguity)}.
\end{itemize}

\begin{table}[t]
\centering
\vspace{-0.1in}
\caption{Experimental setup of repairing methods}
\vspace{-0.15in}
\label{table:Effectiveness Setup}
\resizebox{0.7\columnwidth}{!}{%
\begin{tabular}{c|c|cl|c}
\hline
Scheme    & Prototype & \multicolumn{2}{c|}{Error identification} & Supplementary information             \\ \hline
Rule-Exe  & MAC-SQL   & \multicolumn{2}{c|}{Subset of queries}    & Error Message                         \\ 
LLM-Plain & DIN-SQL   & \multicolumn{2}{c|}{All queries}          & -                                     \\ 
LLM-Exe   & CHESS     & \multicolumn{2}{c|}{All queries}          & Execution Result                      \\ 
LLM-Value & CHESS     & \multicolumn{2}{c|}{All queries}          & Execution Result, Value Specification \\ 
LLM-Extr  & DEA-SQL   & \multicolumn{2}{c|}{All queries}          & Extremum-related Question                     \\ \hline
\end{tabular}%
}
\vspace{-0.15in}
\end{table}

\paragraph{Targets.} For the SQL queries to be repaired, we reuse those generated by MAC-SQL with GPT-3.5 on the \bird benchmark, which contains the richest error types.

\paragraph{Error Analysis.} We use the approach of Section~\ref{sec:emprical_setup_error}.

\begin{table}
    \centering
    \vspace{-0.1in}
    \caption{Repaired queries of MAC-SQL@\bird with GPT-3.5}
    \vspace{-0.15in}
    \label{table:correct_number_repairing}
    \resizebox{0.75\columnwidth}{!}{%
        \begin{tabular}{|l|c|c|c|c|c|}
        \hline
        \multicolumn{1}{|c|}{} & Rule-Exe          & LLM-Plain           & LLM-Exe           & LLM-Value          & LLM-Extr       \\ \hline
        Repaired queries                 & \,\,75 & \,140 & \,148 & \,\,\,161 & \,137 \\ \hline
        Mis-repaired queries                 & \,\,\,\,4  & \,\,\,47  & \,\,\,49  & \,\,\,\,\,53  & \,\,\,55 \\ \hline
        \textbf{Overall Improvement}                          & +71 & +93 & +99 & +108 & +82 \\ \hline
        \end{tabular}%
    }
    \vspace{-0.2in}
\end{table}

\subsection{ICL-Based Repairing Result}

\subsubsection{Correctness Improvement.}\hfill

The performance of each scheme is shown in the right part of Table~\ref{table: error result}. 
\emph{Rule-Exe}, \emph{LLM-Exe}, and \emph{LLM-Value} successfully repair 52-70\% syntax and schema errors. In contrast, \emph{LLM-Plain} and \emph{LLM-Extr} only fix 37.5\% and 38.5\% SQL queries containing these errors, due to lack of DBMS feedback. It highlights the significant impact of incorporating DBMS feedback in LLM inference process.

However, these methods only resolve less than half of logic and convention errors. 
Particularly, most \emph{implicit type convention} errors remain after the repairing. For the \emph{violating value specification} errors, \emph{LLM-Value} fixes 44\% of them, while other schemes fix 24-32\%. It demonstrates that providing targeted information in prompt is helpful for ICL-based repairing.

All five methods face trouble in tackling semantic errors. Most methods repair 23.7-27.3\% of them, while \emph{Rule-Exe} only repairs 9.5\%. In addition, most unfocused errors remain, or even introduced more, after the repairing. Error types of other categories, like \emph{implicit type convention} and \emph{ascending sort with NULL}, also share a similar phenomenon.
It shows the inner incapability of ICL-based methods to understand and repair some types of errors.

\begin{tcolorbox}[width=\linewidth, boxrule=0pt, sharp corners=all,
 left=2pt, right=2pt, top=2pt, bottom=2pt, colback=gray!20]
\textbf{Finding 5}: Rule-based repairing method detects 47.0\% incorrect SQL queries, but only successfully repairs 23.1\% of them (with 5.3\% mis-repair) due to lacking understanding of the errors. 
\end{tcolorbox}

\begin{tcolorbox}[width=\linewidth, boxrule=0pt, sharp corners=all,
 left=2pt, right=2pt, top=2pt, bottom=2pt, colback=gray!20]
\textbf{Finding 6}:
The execution information helps ICL-based techniques repair 14.3\% incorrect queries, and the value specifications further help reduce 1.4\% more. It highlights the benefits brought from extra information. 
\end{tcolorbox}

\subsubsection{Incorrect Repairs.}\hfill

As shown in Table~\ref{table:correct_number_repairing}, not all repairing attempts succeed. In fact, there are three types of incorrect repairing: (1) only repairing a subset of errors; (2) transforming one type of error into another; and (3) introducing errors into a correct SQL query (\ie, mis-repair).

\begin{wrapfigure}{r}{0.4\linewidth}
    \centering
    \includegraphics[width=\linewidth]{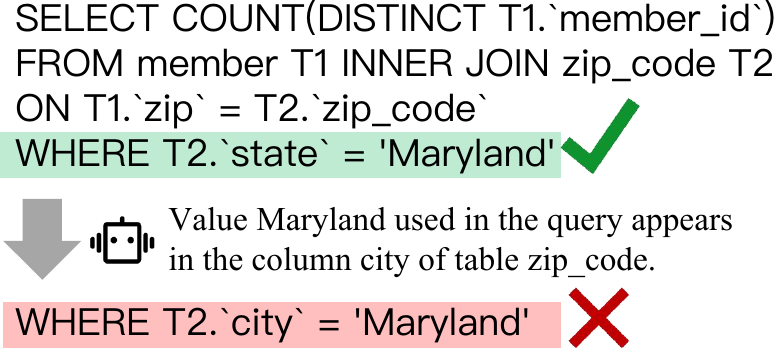}\
    \vspace{-0.3in}
    \caption{
        Mis-repairing \footnotesize{\textnormal{(\#1373@\bird in \emph{LLM-Value} with GPT-3.5)}}.
    }
    \vspace{-0.1in}
    \label{fig:False Repairing}
\end{wrapfigure}
\begin{figure}
    \centering
    \vspace{-0.1in}
        \includegraphics[width=\linewidth]{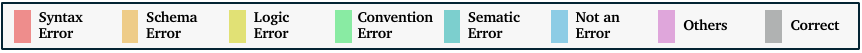}\\
        \includegraphics[width=0.115\linewidth]{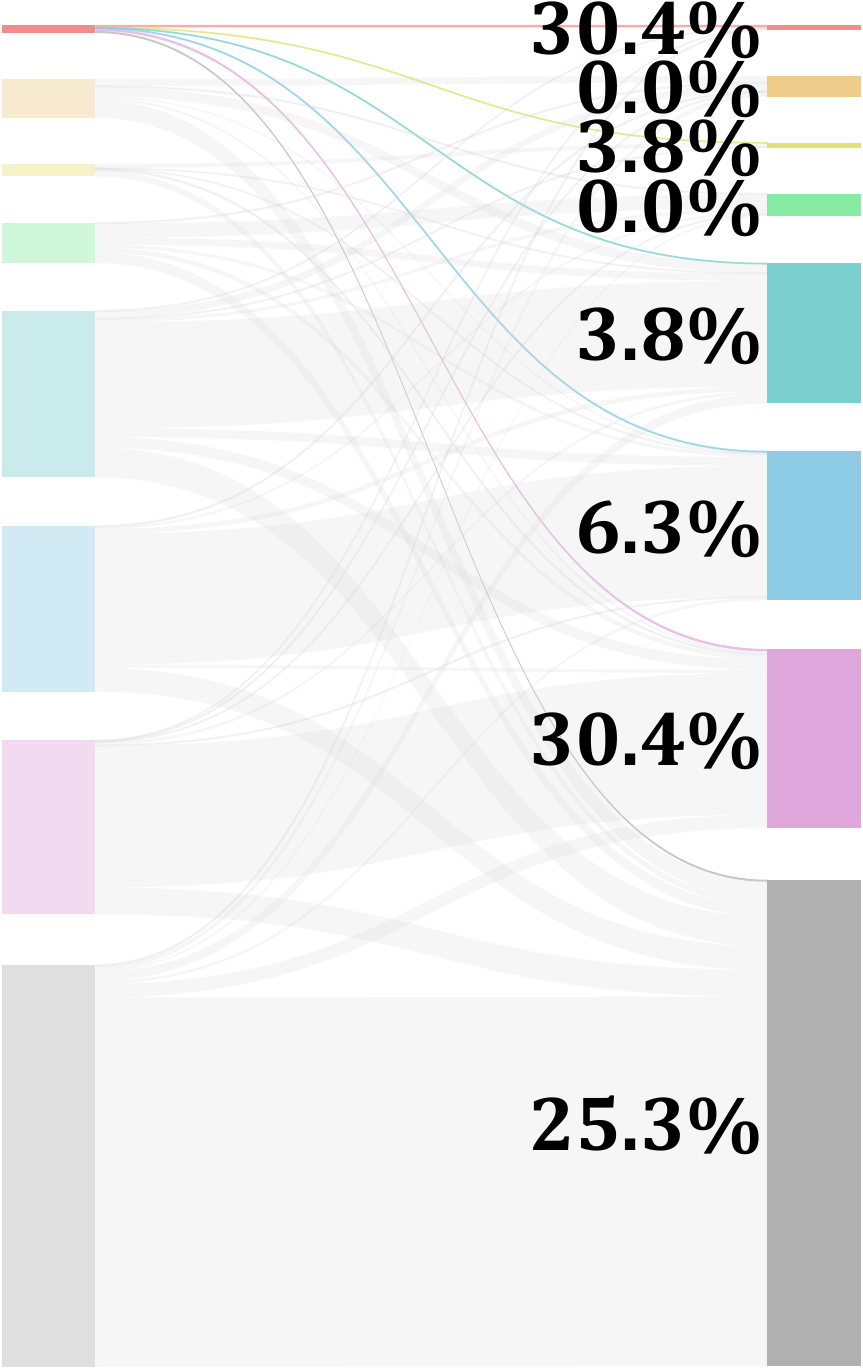}
        \includegraphics[width=0.115\linewidth]{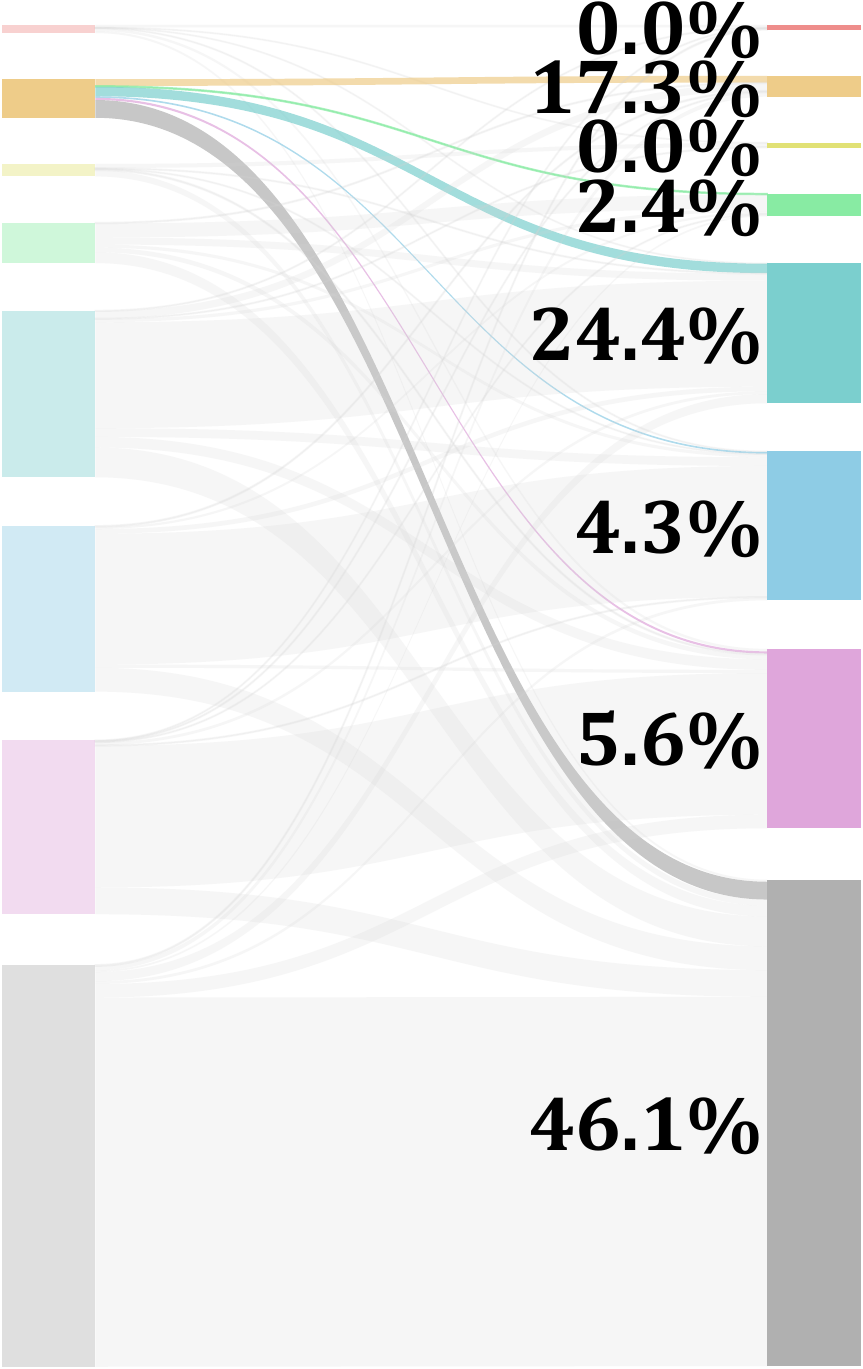}
        \includegraphics[width=0.115\linewidth]{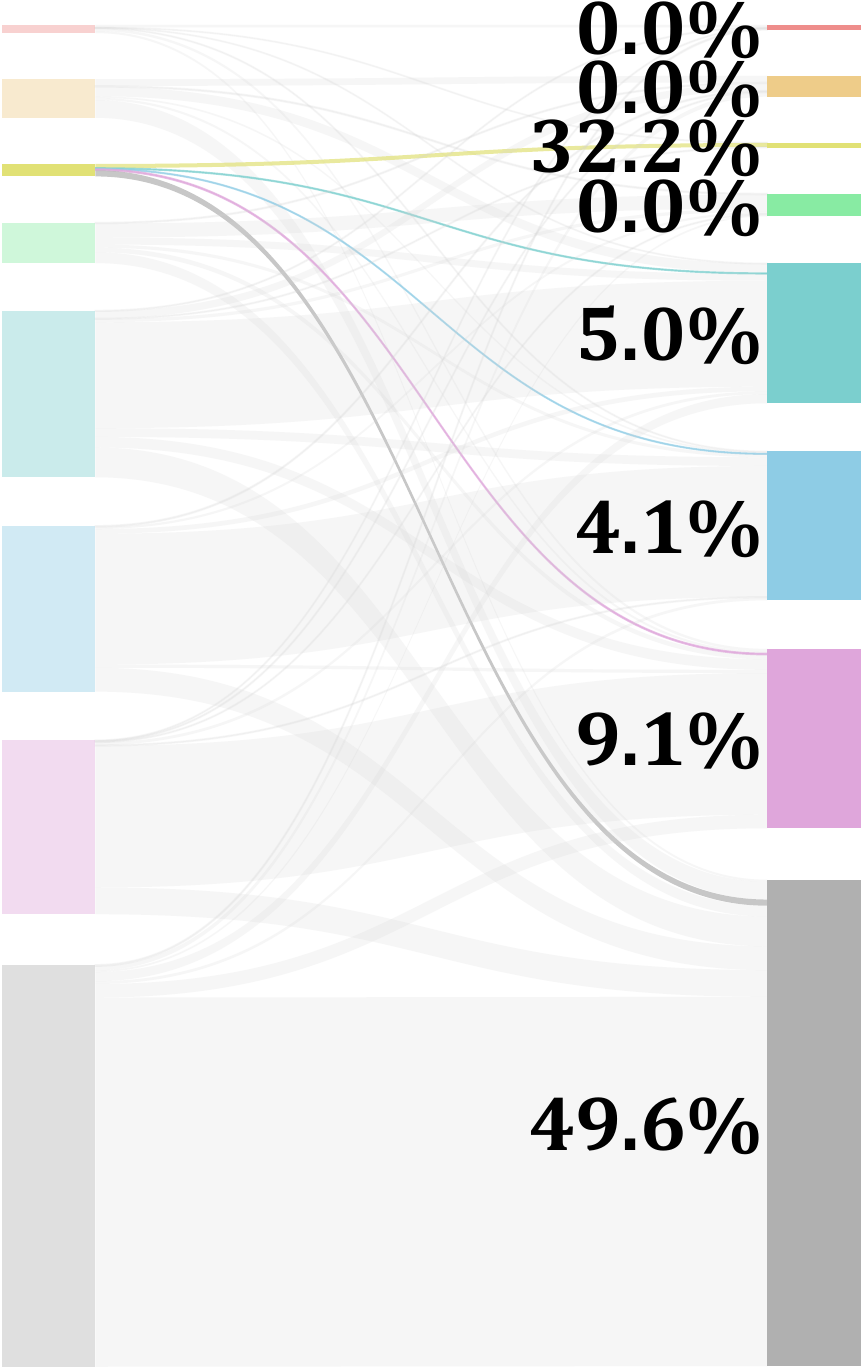}
        \includegraphics[width=0.115\linewidth]{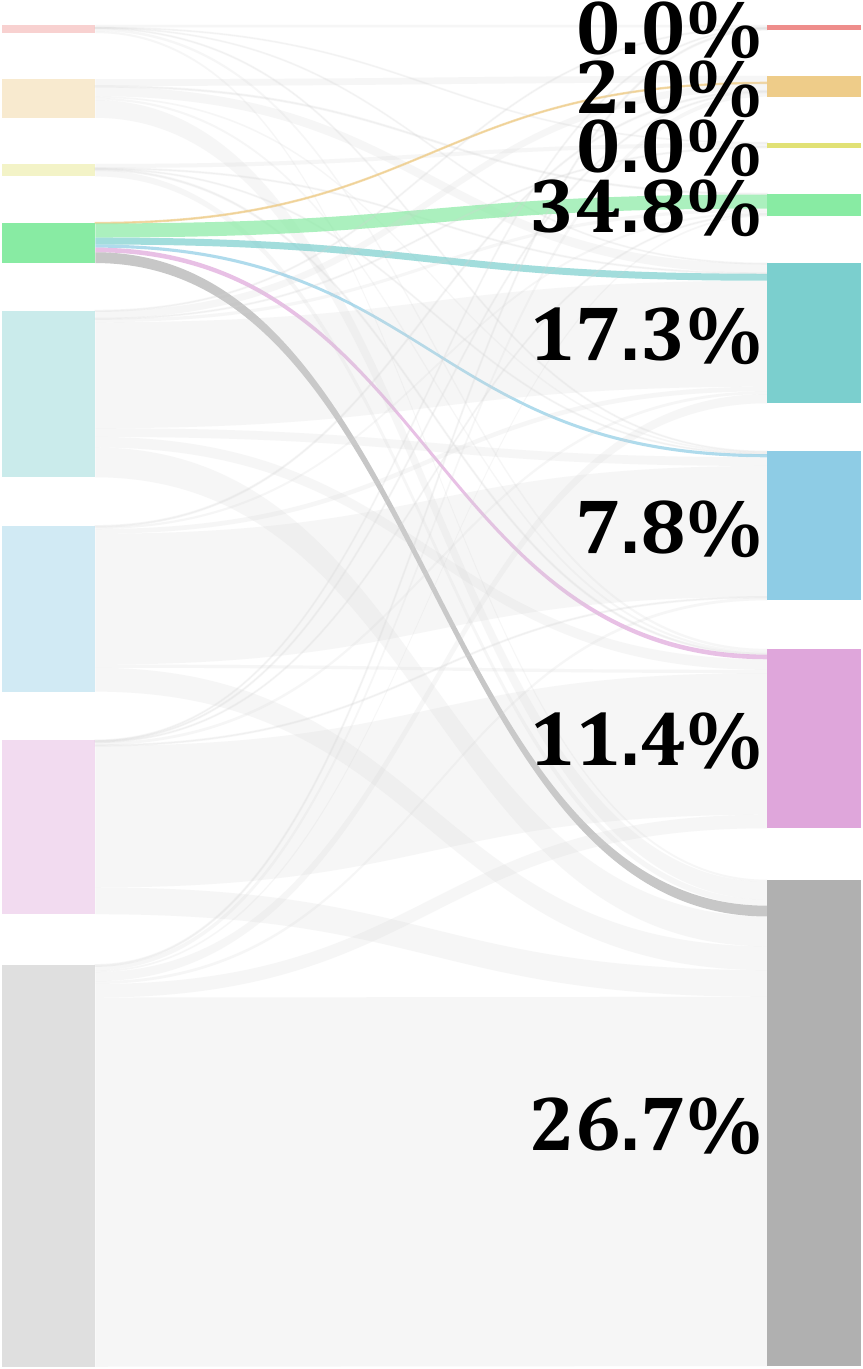}
        \includegraphics[width=0.115\linewidth]{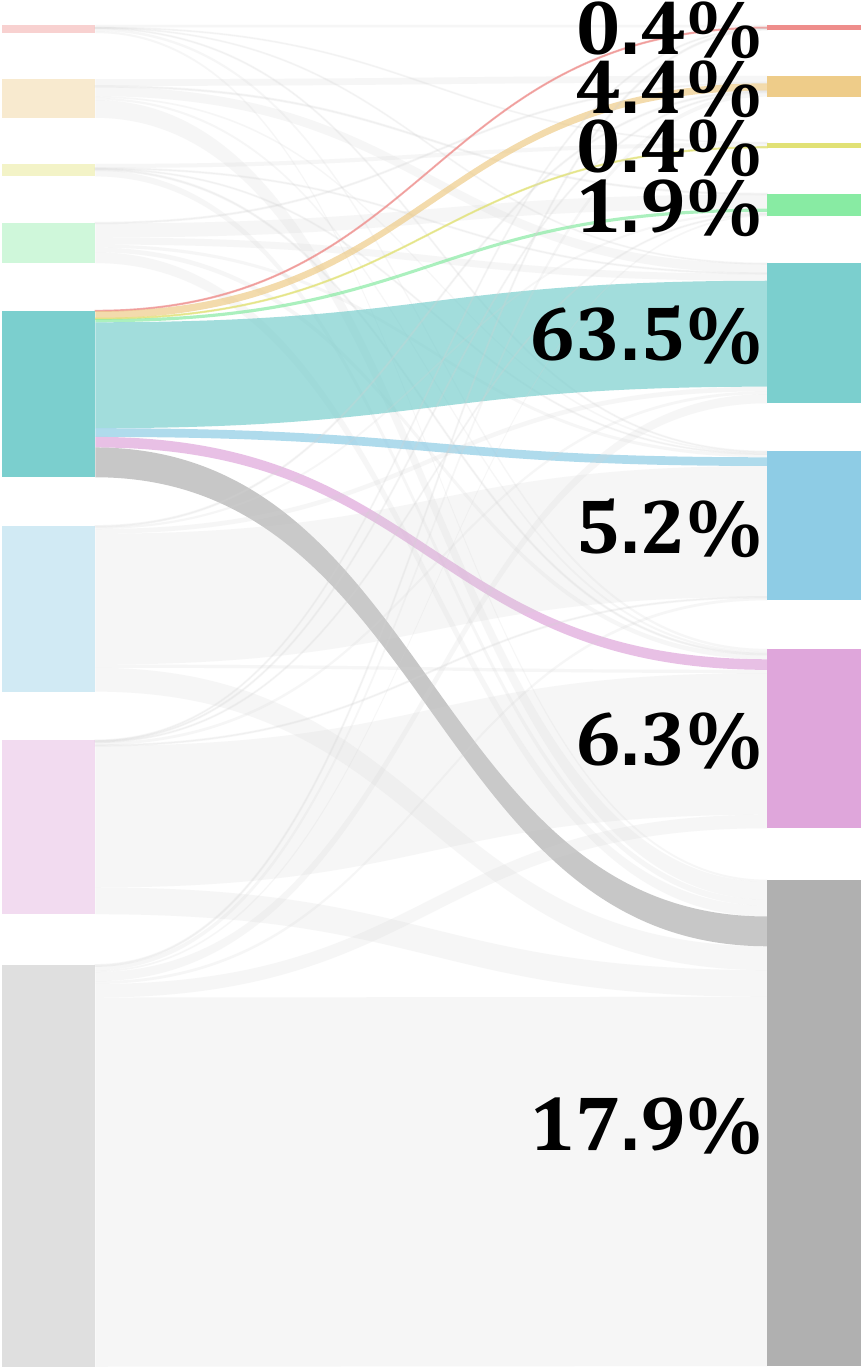}
        \includegraphics[width=0.115\linewidth]{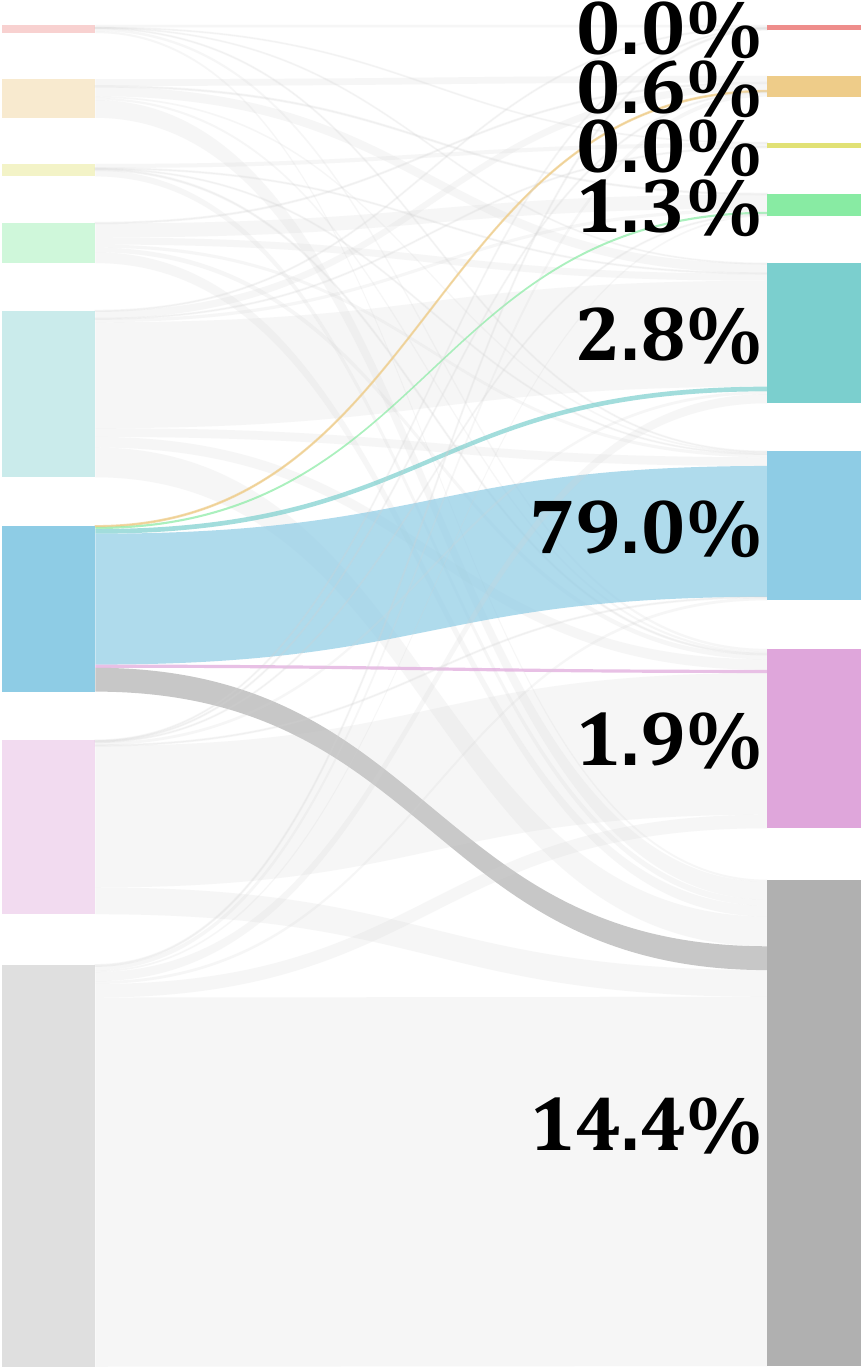}
        \includegraphics[width=0.115\linewidth]{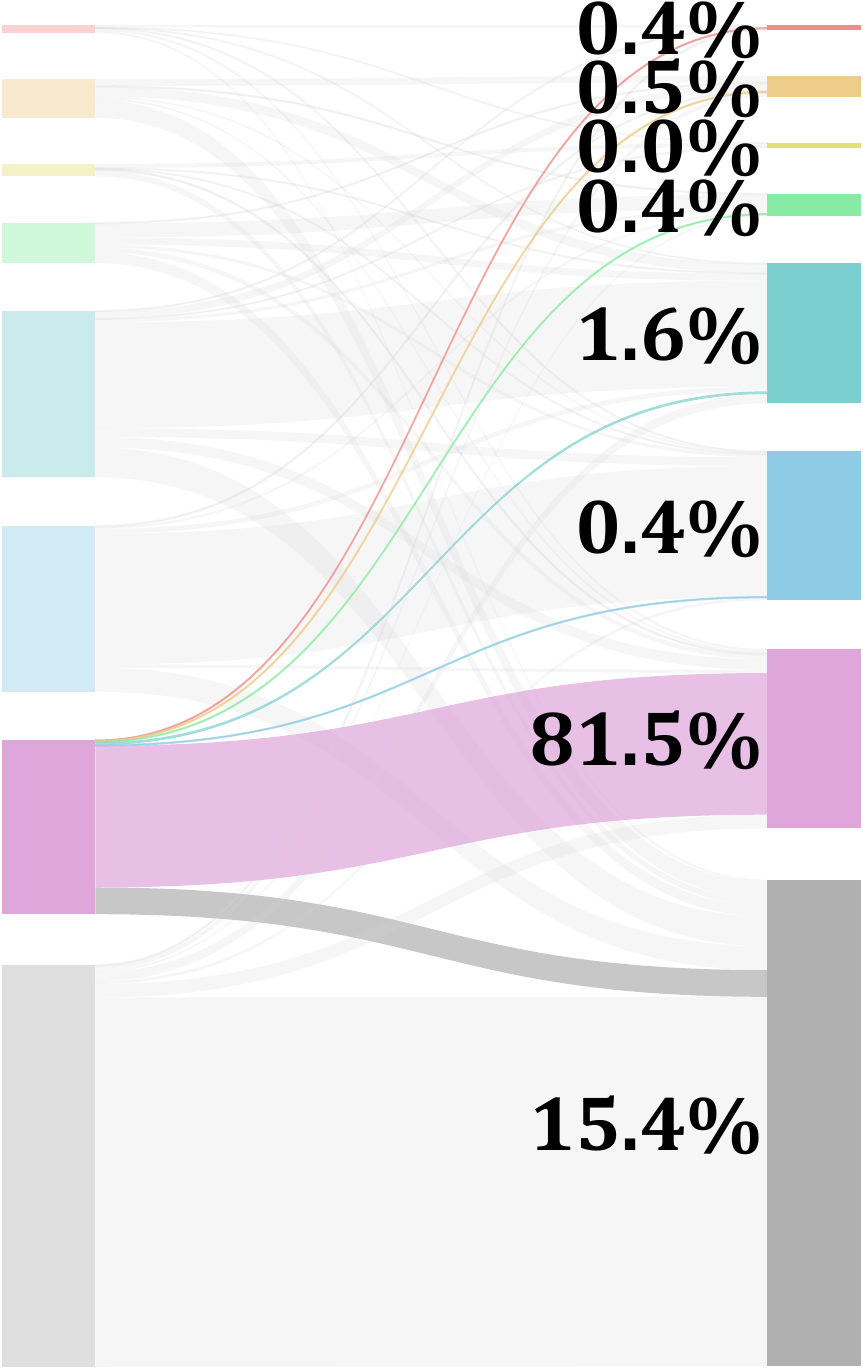}
        \includegraphics[width=0.115\linewidth]{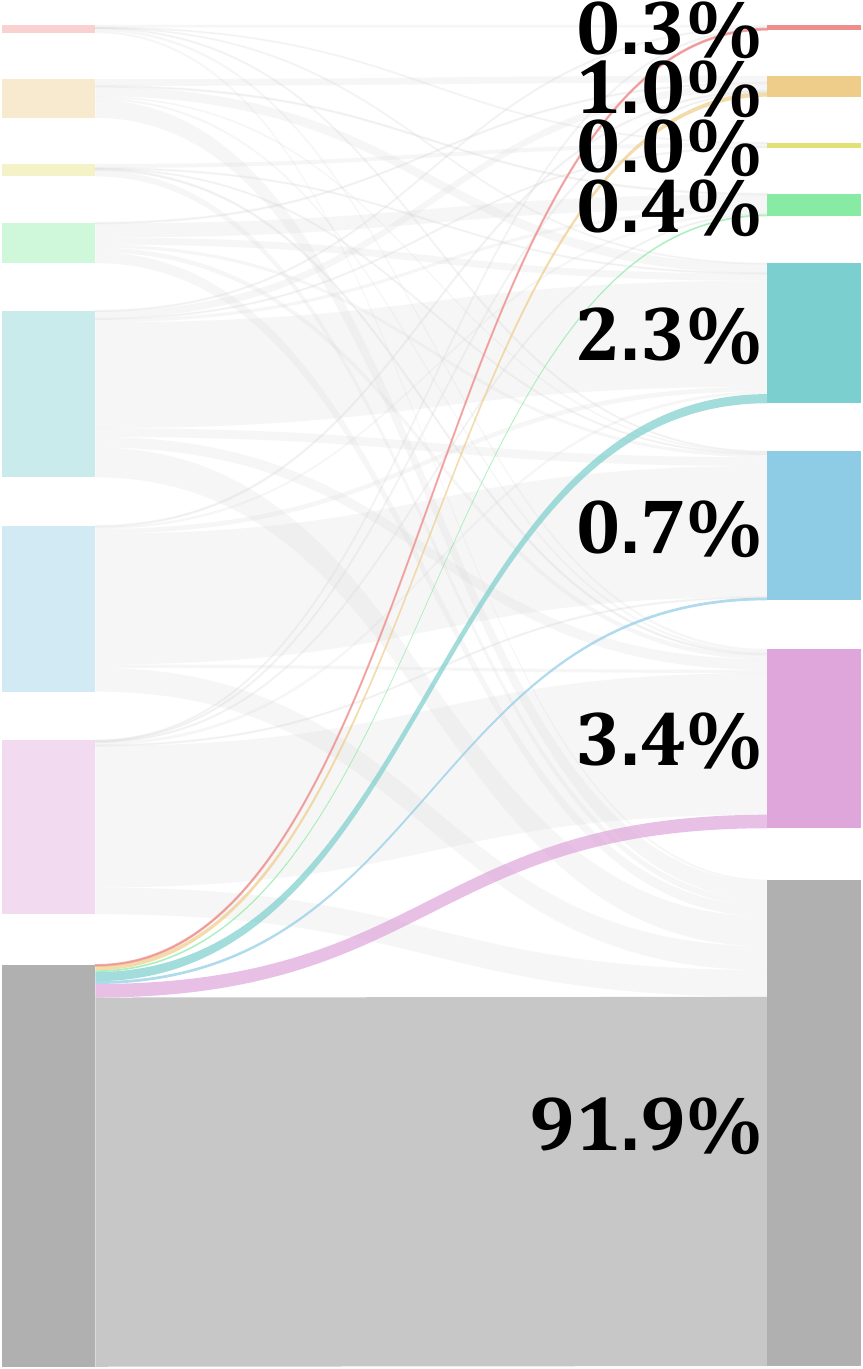}
    \vspace{-0.1in}
    \caption{
        \emph{LLM-Value}'s repairing result @\bird with GPT-3.5 
    }
    \vspace{-0.1in}
    \label{fig:ErrorType Flow(ICL-Based)}
\end{figure}

Figure~\ref{fig:ErrorType Flow(ICL-Based)} details the repairing results of \emph{LLM-Value}. It is effective in repairing intent-independent errors. 
As it regenerates all queries without examining their correctness, it introduces errors into 8.1\% of the correct SQL queries, including 2.3\% into semantic errors and 3.4\% into others category. Note that, these two types of errors are also extremely hard to repair, of which the majority remain after repairing. Figure~\ref{fig:False Repairing} shows an incorrect repair of \emph{LLM-Value}, where \emph{LLM-Value} incorrectly restricts \code{``Maryland''} to the \code{city} column.

We observe similar trends on other repairing schemes except \emph{Rule-Exe} --- on average, they repair 147 incorrect queries and meanwhile introduce 51 more erroneous ones. 
It implies that simple errors (\ie, syntax, schema, logic, and convention errors), if mis-repaired, are often transformed into complex ones that are hard to detect and repair (\ie, semantic errors and others). That is, a mis-repair would exacerbate the errors! We even observe an increment of incorrect SQL queries, after applying \emph{LLM-Extr}.

In contrast, \emph{Rule-Exe} only repairs queries that are likely to be incorrect, repairing 78 SQL queries with only 4 mis-repairs. It has 72.1\% less overhead due to fewer LLM invocations. While efficient, it has the smallest correctness improvement due to insufficient understanding of error types.

\begin{tcolorbox}[width=\linewidth, boxrule=0pt, sharp corners=all,
 left=2pt, right=2pt, top=2pt, bottom=2pt, colback=gray!20]
\textbf{Finding 7}: 
Poorly designed repairing techniques, especially those asking LLMs to re-generate all queries, eventually introduce or exacerbate errors in SQL queries.
\end{tcolorbox}

%% file: sections/5-Tool.tex
\label{sec:tool}

\begin{figure}[t]
    \centering
    \includegraphics[width=1\linewidth]{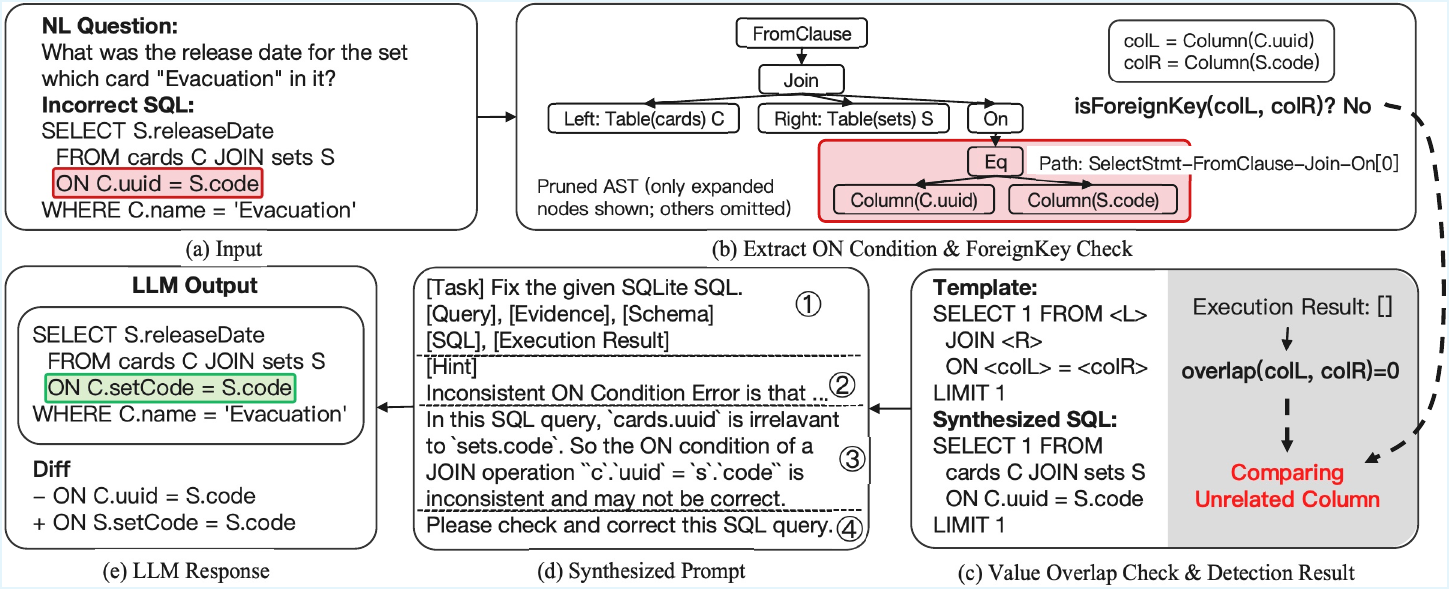}
    \vspace{-0.2in}
    \caption{\edit{A running example of \tool}}
    \label{fig:running_example}
    \vspace{-0.2in}
\end{figure}

\subsection{\tool Design}

We propose \tool, an automated error detection and repairing framework built upon our empirical findings (Section~\ref{sec:errors}\&\ref{sec:method_effectiveness}). Generally, \tool first ensures SQL query executability by repairing syntax and schema errors. Once the query is executable, \tool repairs the remaining errors. If the query remains non-executable, \tool terminates the repair process and returns the original query. Note that, a mis-report of error may be resolved during the LLM-repairing stage.

Driven by the anti-patterns from error taxonomy(Finding~1--3), \tool employs a universal \emph{rule-based detector} to efficiently identify errors, including DBMS error-message checks, AST anti-pattern checks, schema checks, and data/constrain probes. Semantic errors are detected by examining query results, as they lack clear anti-patterns (Section~\ref{sec:consequences_error}).

For repairing, we implement a hybrid, \emph{rule-first strategy} to balance effectiveness and overhead. Unlike prior works that rely solely on limited rules (Finding~5) or risky ``repair-all'' LLM generations (Finding~7), \tool first attempts deterministic rule-based repairs. It invokes the LLM-based repairer only as a \emph{fallback} when rule-based solutions are inapplicable or involve ambiguity.
Guided by Finding~6, we design \emph{error-specific prompts}, to minimize hallucination by providing the precise context needed for accurate repair, including (1) relevant schema metadata, (2) execution results/error messages, (3) error descriptions, (4) error locations, and (5) repair guidelines.

\input{tables/chapter_5_rules_table}

\tool supports all types of execution failures, logic errors, and convention errors, except for \emph{missing quote}.  \tool's modular design allows easy extension of repairing methods. Table~\ref{tab:detailed_rule} summarizes its detection and repair strategies, which are directly derived from the anti-patterns identified in Section~\ref{sec:errors}.

% \vspace{3pt}
\subsection{A Running Example.}

\edit{
Figure~\ref{fig:running_example} shows an illustrative example of \tool workflow. The initial SQL query incorrectly joins \texttt{cards.uuid} and \texttt{sets.code}, which are semantically unrelated.

\noindent\textbf{Step 1: Rule-based Detection.} \tool follows Algorithm~\ref{alg:detection_unrelated} to detect \textit{Comparing Unrelated Column} error. Since these columns lack a foreign key relationship, \tool reports error after confirming no value overlap by executing a lightweight \textit{probe query} (Figure ~\ref{fig:running_example}(c)).

\noindent\textbf{Step 2: Prompt Construction.}
\tool constructs a template-based prompt (Figure~\ref{fig:running_example}(d)) to guide the repair. It  contains
(1) \emph{Context} (\ding{172}), which includes NL question, database schema with foreign keys, and the erroneous SQL query; and
(2) \emph{Hint}, which injects specific guidance derived from the detector, including \textit{Error Description} (\ding{173}), \textit{Error Localization} (\ding{174}), and \textit{Repair Instruction} (\ding{175}).

\noindent\textbf{Step 3: LLM-based Repair.}
Guided by the explicit foreign key context and error hint, the LLM revises the join condition to \texttt{C.setCode = S.code} and successfully resolves the error (Fig.~\ref{fig:running_example}(e)).
}

\input{sections/algo.tex}

\subsection{Evaluation}

\subsubsection{Methodology.} \hfill
\label{sec:eval_method}

\emph{Schemes.} 
We evaluate \tool against the repairing module of MAC-SQL, DIN-SQL, CHESS, and DEA-SQL, by comparing their repairing result of the corresponding generated SQL queries. 
We also compare \tool with the five basic repairing methods in Section~\ref{sec:method_effectiveness}.

To make a fair comparison, in each setting, \tool incorporates the same LLM that generates the SQL queries as its backend.

\emph{Targets.} We use two groups of Text-to-SQL tasks. (a) We adopt the benchmark of Section~\ref{sec:errors}, except for gold errors and \code{flight\_1} database due to their low quality, as discussed in Section~\ref{sec:error_bench_quality}. To ensure consistency, we reuse the generated SQL queries from Section~\ref{sec:errors}; 
(b) To evaluate \tool's generalizability, we randomly select 10 databases with 1,179 tasks from \bird training split (as its test split is closed-source) and 20 databases with 1,032 tasks from \spider test split, which are not included in our earlier study. We further adopt the dev and seed splits (599 text-to-SQL tasks) of \sciencebenchmark~\cite{scibench}, a most recent and challenging Text-to-SQL benchmark with different task types from \bird and \spider. We then use MAC-SQL to generate SQL queries as targets.% for detection and repairing. 

\emph{Hardware.} All experiments were conducted on a 64-bit CentOS-7 machine with a 128-core CPU at \SI{2.9}{GHz}, \SI{512}{GB} RAM, and 8 RTX4090 \SI{24}{GB} memory GPUs.

\subsubsection{Error Detection Capability of \tool.}

\edit{The existing repair solutions do not perform taxonomy-aware detection. Therefore, we only reports \tool's detection results broken down by the error taxonomy proposed in Section~\ref{sec:errors}.}

The result is summarized in Table~\ref{table: tool_detection_result}. Overall, \tool successfully detects 1428 errors with only 179 false detections and 1416 SQL queries with only 265 false detections. Note that, 56.3\% of the detected errors do not have easy-observable symptoms like execution failure.

\input{tables/chapter_5_detection_result}

\subsubsection{Error Repair Capability of \tool.}
Table~\ref{tab:repair_summary_c5} \edit{compares the repairing results of MapleDoctor against the original repair pipelines of each technique}. Overall, \tool successfully repairs 727 SQL queries with only introducing 75 errors by mis-repairs.
Note that, \tool only has 9.4\% mis-repairs on GPT-3.5 generated SQL queries, showing its effectiveness in understanding the text-to-SQL errors. 
Comparatively, the baselines only achieve 0-74.6\% correctness improvement of \tool, while DIN-SQL and DEA-SQL even introduce more errors than they repair (details in Section~\ref{sec:method_effectiveness}). They either only cast looks to error consequences or simply repair all the queries.

\begin{table*}
\caption{\edit{Error repairing results: \tool vs. original repair pipelines.}}
\vspace{-0.15in}
\label{tab:repair_summary_c5}

\small
\resizebox{0.6\linewidth}{!}{%
\edit{
\begin{threeparttable}[b]
\begin{tabular}{ll|rrr|rrr|r}
\toprule
& & \multicolumn{3}{c|}{\textbf{Original Pipeline}} 
& \multicolumn{3}{c|}{\textbf{\tool}} 
& \\
\textbf{Benchmark} & \textbf{Technique@LLM} 
& Rep. & Mis. & Net 
& Rep. & Mis. & Net 
& $\Delta$ \\
\midrule
\multirow{4}{*}{\bird}
& MAC-SQL @ GPT-3.5  &  75 &   6 &  +69 & 131 &   8 & +123 & \textbf{+54} \\
& DIN-SQL @ GPT-3.5  &  61 & 113 &  -52 & 118 &  15 & +103 & \textbf{+155} \\
& CHESS @ GPT-3.5    & 127 &  11 & +116 & 151 &  11 & +140 & \textbf{+24} \\
& MAC-SQL @ GPT-4o   &  24 &   1 &  +23 &  65 &  13 &  +52 & \textbf{+29} \\
\midrule
\multirow{6}{*}{\spider}
& MAC-SQL @ GPT-3.5  &  11 &   0 &  +11 &  46 &   2 &  +44 & \textbf{+33} \\
& DIN-SQL @ GPT-3.5  &  38 &  37 &   +1 &  66 &   2 &  +64 & \textbf{+63} \\
& CHESS @ GPT-3.5    &  55 &   2 &  +53 &  76 &   4 &  +72 & \textbf{+19} \\
& DEA-SQL @ GPT-3.5  &  25 &  64 &  -39 &  34 &   4 &  +30 & \textbf{+69} \\
& MAC-SQL @ GPT-4o   &   0 &   0 &    0 &  29 &  10 &  +19 & \textbf{+19} \\
& DEA-SQL @ GPT-4o   &   8 &  16 &   -8 &  11 &   6 &   +5 & \textbf{+13} \\
\midrule
\multicolumn{2}{l|}{\textbf{Total}} 
& 424 & 250 & +174 & 727 &  75 & +652 & \textbf{+478} \\
\bottomrule
\end{tabular}
\begin{tablenotes}
\small
\item [1] Rep. = Repair; Mis. = Mis-Repair; Net = Rep. - Mis.; $\Delta$ = \tool Net $-$ Original Net. 
\end{tablenotes}
\end{threeparttable}%
}
}
\vspace{-0.2in}
\end{table*}

We further look into the intent-independent errors fixed by \tool and baselines, taking the same setting as Section~\ref{sec:method_effectiveness}, as \tool's main focus is intent-independent errors. As shown in table~\ref{tab:detail_repair}, \edit{\tool eliminates 217 out of 293 intent-independent errors (74.1\% reduction), while the basic repairing methods only resolve 87 to 123 errors (29.7-42.0\%).}

Particularly, \tool repairs most syntax errors and all logic errors, benefiting from detailed error descriptions and repairing instructions.
Among the 293 detected errors in Table~\ref{tab:detail_repair}, 136 of them are repaired by \tool's rule-based solutions, and the remaining are with LLM invocations.

While having significant correctness improvement, \tool cannot fix all SQL queries, as its rule-based detection solutions only cover a subset of errors that have specific symptoms. In addition, a query becomes correct only when all its errors are successfully repaired, which is a non-trivial task.

\edit{
\subsubsection{Analysis of Mis-repairs.}
\label{sec:tool_misrepairs}

We analyze all 75 mis-repairs produced by \tool. Most of them (72/75) involve the LLM-based repairer: a detector flags a false error, \tool invokes the LLM with the corresponding hints, and the generated rewrite may change the query semantics.
The remaining mis-repairs (3/75) arise from rule-based rewrites, suggesting that the rule-based repair only occasionally alter semantics. 
Table~\ref{tab:misrepair_patterns} summarizes typical mis-repair patterns, their triggering detectors/repairers, and a representative example for each pattern.

\begin{table}
\centering
\caption{Intent-independent errors before and after repair of all schemes \footnotesize{\textnormal{(MAC-SQL@\bird with GPT-3.5)}}.}
\vspace{-0.1in}
\label{tab:detail_repair}
\setlength{\tabcolsep}{2pt}
\resizebox{0.75\columnwidth}{!}{%
\begin{threeparttable}[b]
\begin{tabular}{|l|c|c|c|c|c|c|c|}
\hline
\multicolumn{1}{|c|}{}              & Before       & Rule-Exe     & LLM-Plain    & LLM-Exe      & LLM-Value    & LLM-Extr     & \textbf{\tool} \\ \hline
Function Hallucination              & \,\,\,\,\,\,6            & \,\,\,\,\,\,0            & \,\,\,\,\,\,3            & \,\,\,\,\,\,1            & \,\,\,\,\,\,1            & \,\,\,\,\,\,4            & \,\,\,\,\,\,1            \\ \hline
\rowcolor[HTML]{D9D9D9} 
\textbf{SUM: Syntax Error}          & \textbf{\,\,\,\,\,\,6}   & \textbf{\,\,\,\,\,\,0}   & \textbf{\,\,\,\,\,\,3}   & \textbf{\,\,\,\,\,\,1}   & \textbf{\,\,\,\,\,\,1}   & \textbf{\,\,\,\,\,\,4}   & \textbf{\,\,\,\,\,\,1}   \\ \hline
Table-Column Mismatch               & 144          & \,\,\,\,65           & 107          & \,\,\,\,87           & \,\,\,\,88           & 103          & \,\,\,\,39           \\ \hline
Alias Not Use                       & \,\,\,\,\,\,3            & \,\,\,\,\,\,1            & \,\,\,\,\,\,6            & \,\,\,\,\,\,1            & \,\,\,\,\,\,1            & \,\,\,\,\,\,6            & \,\,\,\,\,\,0            \\ \hline
Guided Missing JOIN                 & \,\,\,\,12           & \,\,\,\,\,\,2            & \,\,\,\,\,\,6            & \,\,\,\,\,\,5            & \,\,\,\,\,\,6            & \,\,\,\,\,\,5            & \,\,\,\,\,\,1            \\ \hline
Schema Hallucination                & \,\,\,\,\,\,0            & \,\,\,\,\,\,1            & \,\,\,\,\,\,0            & \,\,\,\,\,\,1            & \,\,\,\,\,\,1            & \,\,\,\,\,\,0            & \,\,\,\,\,\,0            \\ \hline
\rowcolor[HTML]{D9D9D9} 
\textbf{SUM: Schema Error}          & \textbf{159} & \textbf{\,\,\,\,69}  & \textbf{119} & \textbf{\,\,\,\,94}  & \textbf{\,\,\,\,96}  & \textbf{114} & \textbf{\,\,\,\,40}  \\ \hline
Implicit Type Conversion            & \,\,\,\,12           & \,\,\,\,11           & \,\,\,\,\,\,9            & \,\,\,\,\,\,8            & \,\,\,\,\,\,9            & \,\,\,\,\,\,9            & \,\,\,\,\,\,0            \\ \hline
Using = instead of IN               & \,\,\,\,24           & \,\,\,\,18           & \,\,\,\,\,\,6            & \,\,\,\,\,\,5            & \,\,\,\,\,\,6            & \,\,\,\,\,\,4            & \,\,\,\,\,\,0            \\ \hline
Ascending Sort with NULL            & \,\,\,\,\,\,2            & \,\,\,\,\,\,2            & \,\,\,\,\,\,3            & \,\,\,\,\,\,3            & \,\,\,\,\,\,3            & \,\,\,\,\,\,4            & \,\,\,\,\,\,0            \\ \hline
\rowcolor[HTML]{D9D9D9} 
\textbf{SUM: Logic Error}           & \textbf{\,\,\,\,38}  & \textbf{\,\,\,\,31}  & \textbf{\,\,\,\,18}  & \textbf{\,\,\,\,16}  & \textbf{\,\,\,\,18}  & \textbf{\,\,\,\,17}  & \textbf{\,\,\,\,\,\,0}   \\ \hline
Value Specification                 & \,\,\,\,75           & \,\,\,\,68           & \,\,\,\,60           & \,\,\,\,69           & \,\,\,\,50           & \,\,\,\,59           & \,\,\,\,30           \\ \hline
Comparison Misuse                   & \,\,\,\,12           & \,\,\,\,\,\,9            & \,\,\,\,\,\,5            & \,\,\,\,\,\,4            & \,\,\,\,\,\,4            & \,\,\,\,\,\,6            & \,\,\,\,\,\,4            \\ \hline
Comparing Unrelated Columns         & \,\,\,\,\,\,3            & \,\,\,\,\,\,2            & \,\,\,\,\,\,1            & \,\,\,\,\,\,1            & \,\,\,\,\,\,1            & \,\,\,\,\,\,2            & \,\,\,\,\,\,0            \\ \hline
\rowcolor[HTML]{D9D9D9} 
\textbf{SUM: Convention Error}      & \textbf{\,\,\,\,90}  & \textbf{\,\,\,\,79}  & \textbf{\,\,\,\,66}  & \textbf{\,\,\,\,74}  & \textbf{\,\,\,\,55}  & \textbf{\,\,\,\,67}  & \textbf{\,\,\,\,34}  \\ \hline
\rowcolor[HTML]{A6A6A6} 
\textbf{Total number of all errors} & \textbf{293} & \textbf{179} & \textbf{206} & \textbf{185} & \textbf{170} & \textbf{202} & \textbf{\,\,\,\,75}  \\ \hline
\end{tabular}
\begin{tablenotes}
\item [1] \tool may detect and repair errors out of the ``edit path to correctness''. Therefore, it has different error numbers in \emph{Before} column from Table~\ref{table: error result}.
\end{tablenotes}
\end{threeparttable}%
}
\vspace{-0.1in}
\end{table}

\begin{table}
\centering
\caption{\edit{Typical mis-repair patterns in \tool: triggers and representative examples.}}
\vspace{-0.1in}
\label{tab:misrepair_patterns}
\scriptsize
\setlength{\tabcolsep}{3pt}
\resizebox{\linewidth}{!}{%
\begin{tabular}{|l|c|l|l|}
\hline
\textbf{Pattern} & \textbf{\#} & \textbf{Trigger (detector/repairer)} & \textbf{Example (one case)} \\
\hline
Extremum rewrite ($\texttt{MAX/MIN}\rightarrow \texttt{ORDER BY+LIMIT 1}$) & 20 & Extremum-pattern detector (MAX/MIN-in-subquery) $\Rightarrow$ LLM repair & CHESS \bird GPT-3.5 \#621 \\
\hline
CAST injection (\texttt{CAST}/\texttt{REPLACE+CAST} on text-like values) & 17 & \emph{Aggregation/Comparison Misuse} detector $\Rightarrow$ LLM repair & CHESS \bird GPT-3.5 \#879 \\
\hline
Value over-correction (replace constants with ``recommended'' values) & 9 & \emph{Value Specification} detector $\Rightarrow$ LLM repair & DIN-SQL \bird GPT-3.5 \#451 \\
\hline
Value over-correction from empty/NULL symptoms & 2 & Suspicious-symptom detector (empty/NULL) $\Rightarrow$ LLM repair & DIN-SQL \bird GPT-3.5 \#74 \\
\hline
Projection drop (drop a requested output column) & 9 & Projection-consistency detector (NL output $\leftrightarrow$ SELECT list) $\Rightarrow$ LLM repair & DEA-SQL \spider GPT-4o \#565 \\
\hline
Schema/column hallucination (introduce non-existent columns/tables) & 3 & Ambiguous repair after \emph{Value Spec./Comparison Misuse} hints $\Rightarrow$ LLM repair & CHESS \bird GPT-3.5 \#74 \\
\hline
Column substitution (replace with a wrong-but-existing column) & 4 & \emph{Aggregation/Comparison Misuse} detector $\Rightarrow$ LLM repair & CHESS \bird GPT-3.5 \#1005 \\
\hline
Predicate flip/contradiction (break negation/set logic) & 2 & \emph{Value Specification} detector $\Rightarrow$ LLM repair & DIN-SQL \bird GPT-3.5 \#529 \\
\hline
Rule-based over-triggering (deterministic rewrite changes semantics) & 3 & \emph{Using = instead of IN} / column-spell repairers (rule-based) & MAC-SQL \bird GPT-3.5 \#1284 \\
\hline
\end{tabular}%
}
\vspace{-0.1in}
\end{table}
}

\subsubsection{Generalizability of \tool.}

We further evaluate the performance of \tool on different benchmarks and queries generated by various LLMs.

\emph{For benchmarks}, we randomly select 10 databases with 1,179 tasks from \bird training split (as its test split is closed-source) and 20 databases with 1,032 tasks from \spider test split, which are not included in our earlier study. We further adopt the dev and seed splits (599 text-to-SQL tasks) of \sciencebenchmark, a most recent and challenging Text-to-SQL benchmark with different task types from \bird and \spider.

\emph{For LLMs}, we use different LLM backends for MAC-SQL to obtain different sets of SQL queries. We adopt (1) 1 close-source LLM, GPT-3.5-Turbo (GPT3.5); and (2) 4 open-source LLMs, Qwen3-14B (Qwen3), Mistral-Nemo-12B (Mistral), CodeLLaMA-13B (LLaMA) and DeepSeekCoderV2-15B (DSCoder).  

Table~\ref{table: generalizability detection}\&\ref{table: generalizability repairing} summarizes the error detection and repairing results. \tool demonstrates strong generalizability and effectiveness across all benchmarks and LLM settings, outperforming the baselines in nearly all cases by providing a high number of correct repairs (up to 10.1\%) with minimal mis-repairs (less than 0.2\%).

\input{tables/chapter_5_generalizability_detection}

\input{tables/chapter_5_generalizability_repairing}

\subsubsection{Performance Overhead.} Due to computation workload and network transmission, the run-time overhead of repairing solutions is mainly caused by extra LLM invocations. On average, \tool takes \SI{1.2}{s} to examine and repair a SQL query, among which LLM invocation takes \SI{1.1}{s} and database execution takes \SI{0.05}{ms}. In comparison, the LLM-based basic repairing methods take 3.5-\SI{8.0}{s}, and the rule-based method takes \SI{1.0}{s}.

%% file: tables/chapter_5_rules_table.tex
\begin{table}[t]
\centering
\vspace{-0.1in}
\caption{\tool's detection and repairing strategy for each type of error.}
\vspace{-0.1in}
\label{tab:detailed_rule}
\resizebox{\linewidth}{!}{%
\begin{tabular}{|l|p{8cm}|p{7cm}|}
\hline
\textbf{Error Type} & \textbf{Detection Solution} & \textbf{Repair Strategy} \\
\hline
\hline
\multicolumn{3}{|c|}{\edit{\textbf{Execution failures}}} \\ \hline
\emph{Function Hallucination} & Check DBMS error ``no such function''. & Replace with a predefined, supported alternative. \\
\hline
\emph{Table-Column Mismatch} & Check DBMS error ``no such column'', and then check the existence of column in database. & Replace the table with the one that contains such column. If multiple tables fit, it invokes an LLM to judge. \\
\hline
\emph{Unused Alias} & Check DBMS error ``no such column'', and then check whether the table contains the column but the query refers to original table name instead of its alias. & Replace the table name with its alias. \\
\hline
\emph{Schema Hallucination} & Check DBMS error ``no such column'', and check the non-existence of column in database. & Try to find a column with similar name using edit distance. If it fails, it invokes an LLM to judge. \\
\hline
\emph{Missing JOIN} & Check DBMS error ``no such column'', and check whether the column exists in an unjoined table. & Join the missing table using its foreign key. If no foreign keys exist or multiple candidates are found, it invokes an LLM to judge. \\
\hline
\hline
\multicolumn{3}{|c|}{\edit{\textbf{Logic errors}}} \\ \hline
\emph{Implicit Type Conversion} & Find division nodes in AST that have two integer operands without \texttt{CAST} operation. & Cast one of the operands to a float. \\
\hline
\emph{Using = instead of IN} & Execute subqueries to find those with multiple results, and examine the existence of  \texttt{=} operator in join clause. & Replace the \texttt{=} operator with \texttt{IN} operator. \\
\hline
\emph{Ascending Sort with NULL} & Find columns with ascending sorts or \texttt{MIN} aggregation operations, and then check the existence of \texttt{NULL} values. & Add an \texttt{IS NOT NULL} clause to the query. \\
\hline
\hline
\multicolumn{3}{|c|}{\edit{\textbf{Convention errors}}} \\ \hline
\emph{Comparing Unrelated Columns} & Find column pairs that are compared in the query but do not have shared values in the database. & Try to replace one column with a foreign key column of the same table. If multiple foreign key columns fit, invoke an LLM to judge.  \\
% \newline 2. If fails, asks LLM provide the LLM with foreign key information and an error description.
\hline
\emph{Violating Value Specification} & Examine whether the target value violates the column specification. & Invoke an LLM to cluster the target value to the valid values of such column. \\
% Provide the LLM with an error description and a list of valid, similar values from a vector database. \\
\hline
\emph{Aggregation/Comparison Misuse} & Examine whether the target column is sortable. & Invoke an LLM to fix the operator, providing sample values from this column. \\
% Provide the LLM with an error description and sample values from the target column. \\
\hline
\end{tabular}
}
\vspace{-0.2in}
\end{table}

%% file: sections/algo.tex
% ==================== Detection Algorithm ====================
\begin{algorithm}[t]
\caption{Detection of Comparing Unrelated Columns}
\label{alg:detection_unrelated}
% \small
\footnotesize
\DontPrintSemicolon % 不显示行尾分号，显得更干净
\SetKwFunction{FExtract}{GetCols}
\SetKwFunction{FCheckFK}{HasFK}
\SetKwFunction{FExec}{Exec}
\SetInd{0.5em}{0.5em}
\KwIn{SQL Query $Q$, Database Schema $D$}
\KwOut{Boolean $is\_detected$, Set $S$}
$S \leftarrow \emptyset$, $is\_detected \leftarrow \text{False}$\;
$AST \leftarrow \text{Parse}(Q)$\;
\ForEach(\tcp*[f]{Traverse all JOIN clauses}){Join Node $J$ in $AST$}{
    \If{$J$ has condition $C$ of type Equality ($=$)}{
        $L, R \leftarrow \FExtract(C)$ \tcp*[r]{Extract columns from ON condition}
        \If{$L, R$ are valid}{
            \If(\tcp*[f]{Step 1: Check Foreign Key}){\textbf{not} $\FCheckFK(D, L, R)$}{
                $Q_{p} \leftarrow \text{"SELECT 1 FROM } L.Table$ $\text{JOIN } R.Table \text{ ON } L{=}R \text{ LIMIT 1"}$\;
                $Result \leftarrow \FExec(D, Q_{probe})$ \tcp*[r]{Step 2: Check Value Overlap}
                \If{$Result$ is Empty}{
                    $S \leftarrow S \cup \{C\}$, $is\_detected \leftarrow \text{True}$\;
                }
            }
        }
    }
}
\Return{$is\_detected, S$}
\end{algorithm}

%% file: tables/chapter_5_detection_result.tex
\begin{table}
\centering
\caption{Error detection result of \tool. }
\vspace{-0.15in}
\label{table: tool_detection_result}
\resizebox{0.95\columnwidth}{!}{%
\begin{threeparttable}[b]
\label{tab:my-table}
\begin{tabular}{|l|cccc|cccccc|}
\hline
\multicolumn{1}{|c|}{}                              & \multicolumn{4}{c|}{\bird}                                                                                                                                                 & \multicolumn{6}{c|}{\spider}                                                                                                                                                                                                                                                        \\ \cline{2-11} 
\multicolumn{1}{|c|}{}                              & \multicolumn{3}{c|}{GPT-3.5-Turbo}                                                                                                                              & GPT-4o  & \multicolumn{4}{c|}{GPT-3.5-Turbo}                                                                                                                                                                                  & \multicolumn{2}{c|}{GPT-4o}                                  \\ \cline{2-11} 
\multicolumn{1}{|c|}{\multirow{-3}{*}{}}            & \multicolumn{1}{c|}{MAC-SQL}                        & \multicolumn{1}{c|}{DIN-SQL}                        & \multicolumn{1}{c|}{CHESS}                          & MAC-SQL & \multicolumn{1}{c|}{MAC-SQL}                       & \multicolumn{1}{c|}{DIN-SQL}                        & \multicolumn{1}{c|}{CHESS}                          & \multicolumn{1}{c|}{DEA-SQL}                       & \multicolumn{1}{c|}{MAC-SQL}                       & DEA-SQL \\ \hline
Function Hallucination                              & \multicolumn{1}{c|}{6/0}                            & \multicolumn{1}{c|}{1/0}                            & \multicolumn{1}{c|}{18/0}                           & 0/0     & \multicolumn{1}{c|}{0/0}                           & \multicolumn{1}{c|}{0/0}                            & \multicolumn{1}{c|}{0/0}                            & \multicolumn{1}{c|}{0/0}                           & \multicolumn{1}{c|}{0/0}                           & 0/0     \\ \hline
Other Syntax Violation                              & \multicolumn{1}{c|}{105/0}                          & \multicolumn{1}{c|}{64/0}                           & \multicolumn{1}{c|}{61/0}                           & 33/0    & \multicolumn{1}{c|}{13/0}                          & \multicolumn{1}{c|}{49/0}                           & \multicolumn{1}{c|}{44/0}                           & \multicolumn{1}{c|}{20/0}                          & \multicolumn{1}{c|}{2/0}                           & 1/0     \\ \hline
\rowcolor[HTML]{D9D9D9} 
\textbf{SUM: Syntax Error}                          & \multicolumn{1}{c|}{\cellcolor[HTML]{D9D9D9}111/0}  & \multicolumn{1}{c|}{\cellcolor[HTML]{D9D9D9}65/0}   & \multicolumn{1}{c|}{\cellcolor[HTML]{D9D9D9}79/0}   & 33/0    & \multicolumn{1}{c|}{\cellcolor[HTML]{D9D9D9}13/0}  & \multicolumn{1}{c|}{\cellcolor[HTML]{D9D9D9}49/0}   & \multicolumn{1}{c|}{\cellcolor[HTML]{D9D9D9}44/0}   & \multicolumn{1}{c|}{\cellcolor[HTML]{D9D9D9}20/0}  & \multicolumn{1}{c|}{\cellcolor[HTML]{D9D9D9}2/0}   & 1/0     \\ \hline
Table-Column Mismatch                               & \multicolumn{1}{c|}{143/0}                          & \multicolumn{1}{c|}{54/0}                           & \multicolumn{1}{c|}{37/0}                           & 18/0    & \multicolumn{1}{c|}{12/0}                          & \multicolumn{1}{c|}{9/0}                            & \multicolumn{1}{c|}{13/0}                           & \multicolumn{1}{c|}{3/0}                           & \multicolumn{1}{c|}{3/0}                           & 0/0     \\ \hline
Unused Alias                                        & \multicolumn{1}{c|}{3/0}                            & \multicolumn{1}{c|}{0/0}                            & \multicolumn{1}{c|}{18/0}                           & 0/0     & \multicolumn{1}{c|}{6/0}                           & \multicolumn{1}{c|}{18/0}                           & \multicolumn{1}{c|}{2/0}                            & \multicolumn{1}{c|}{0/0}                           & \multicolumn{1}{c|}{0/0}                           & 0/0     \\ \hline
Schema Hallucination                                & \multicolumn{1}{c|}{0/0}                            & \multicolumn{1}{c|}{11/0}                           & \multicolumn{1}{c|}{11/0}                           & 0/0     & \multicolumn{1}{c|}{0/0}                           & \multicolumn{1}{c|}{0/0}                            & \multicolumn{1}{c|}{13/0}                           & \multicolumn{1}{c|}{2/0}                           & \multicolumn{1}{c|}{0/0}                           & 0/0     \\ \hline
Missing JOIN                                        & \multicolumn{1}{c|}{12/0}                           & \multicolumn{1}{c|}{31/0}                           & \multicolumn{1}{c|}{15/0}                           & 6/0     & \multicolumn{1}{c|}{2/0}                           & \multicolumn{1}{c|}{9/0}                            & \multicolumn{1}{c|}{4/0}                            & \multicolumn{1}{c|}{5/0}                           & \multicolumn{1}{c|}{0/0}                           & 0/0     \\ \hline
\rowcolor[HTML]{D9D9D9} 
\textbf{SUM: Schema Error}                          & \multicolumn{1}{c|}{\cellcolor[HTML]{D9D9D9}158/0}  & \multicolumn{1}{c|}{\cellcolor[HTML]{D9D9D9}96/0}   & \multicolumn{1}{c|}{\cellcolor[HTML]{D9D9D9}81/0}   & 24/0    & \multicolumn{1}{c|}{\cellcolor[HTML]{D9D9D9}20/0}  & \multicolumn{1}{c|}{\cellcolor[HTML]{D9D9D9}36/0}   & \multicolumn{1}{c|}{\cellcolor[HTML]{D9D9D9}32/0}   & \multicolumn{1}{c|}{\cellcolor[HTML]{D9D9D9}10/0}  & \multicolumn{1}{c|}{\cellcolor[HTML]{D9D9D9}3/0}   & 0/0     \\ \hline
Implicit Type Conversion                            & \multicolumn{1}{c|}{12/0}                           & \multicolumn{1}{c|}{5/0}                            & \multicolumn{1}{c|}{12/0}                           & 1/0     & \multicolumn{1}{c|}{0/0}                           & \multicolumn{1}{c|}{0/0}                            & \multicolumn{1}{c|}{0/0}                            & \multicolumn{1}{c|}{0/0}                           & \multicolumn{1}{c|}{0/0}                           & 0/0     \\ \hline
Using = instead of IN                               & \multicolumn{1}{c|}{24/9}                           & \multicolumn{1}{c|}{0/0}                            & \multicolumn{1}{c|}{0/1}                            & 15/8    & \multicolumn{1}{c|}{1/0}                           & \multicolumn{1}{c|}{1/0}                            & \multicolumn{1}{c|}{0/0}                            & \multicolumn{1}{c|}{2/0}                           & \multicolumn{1}{c|}{0/0}                           & 1/0     \\ \hline
Ascending Sort with NULL                            & \multicolumn{1}{c|}{2/0}                            & \multicolumn{1}{c|}{1/0}                            & \multicolumn{1}{c|}{2/0}                            & 4/0     & \multicolumn{1}{c|}{0/0}                           & \multicolumn{1}{c|}{0/0}                            & \multicolumn{1}{c|}{1/0}                            & \multicolumn{1}{c|}{0/0}                           & \multicolumn{1}{c|}{0/0}                           & 0/0     \\ \hline
\rowcolor[HTML]{D9D9D9} 
\textbf{SUM: Logic Error}                           & \multicolumn{1}{c|}{\cellcolor[HTML]{D9D9D9}38/9}   & \multicolumn{1}{c|}{\cellcolor[HTML]{D9D9D9}6/0}    & \multicolumn{1}{c|}{\cellcolor[HTML]{D9D9D9}14/1}   & 20/8    & \multicolumn{1}{c|}{\cellcolor[HTML]{D9D9D9}1/0}   & \multicolumn{1}{c|}{\cellcolor[HTML]{D9D9D9}1/0}    & \multicolumn{1}{c|}{\cellcolor[HTML]{D9D9D9}1/0}    & \multicolumn{1}{c|}{\cellcolor[HTML]{D9D9D9}2/0}   & \multicolumn{1}{c|}{\cellcolor[HTML]{D9D9D9}0/0}   & 1/0     \\ \hline
Violating Value Specification                       & \multicolumn{1}{c|}{75/12}                          & \multicolumn{1}{c|}{69/11}                          & \multicolumn{1}{c|}{94/14}                          & 47/13   & \multicolumn{1}{c|}{11/9}                          & \multicolumn{1}{c|}{38/6}                           & \multicolumn{1}{c|}{37/4}                           & \multicolumn{1}{c|}{9/5}                           & \multicolumn{1}{c|}{6/7}                           & 5/5     \\ \hline
Aggregation/Comparison Misuse                       & \multicolumn{1}{c|}{12/5}                           & \multicolumn{1}{c|}{5/10}                           & \multicolumn{1}{c|}{6/8}                            & 3/3     & \multicolumn{1}{c|}{0/3}                           & \multicolumn{1}{c|}{0/3}                            & \multicolumn{1}{c|}{0/2}                            & \multicolumn{1}{c|}{0/3}                           & \multicolumn{1}{c|}{0/1}                           & 0/2     \\ \hline
Comparing Unrelated Columns                         & \multicolumn{1}{c|}{3/0}                            & \multicolumn{1}{c|}{13/0}                           & \multicolumn{1}{c|}{20/2}                           & 1/0     & \multicolumn{1}{c|}{3/0}                           & \multicolumn{1}{c|}{2/1}                            & \multicolumn{1}{c|}{5/2}                            & \multicolumn{1}{c|}{0/0}                           & \multicolumn{1}{c|}{0/0}                           & 0/0     \\ \hline
\rowcolor[HTML]{D9D9D9} 
\textbf{SUM: Convention Error}                      & \multicolumn{1}{c|}{\cellcolor[HTML]{D9D9D9}90/17}  & \multicolumn{1}{c|}{\cellcolor[HTML]{D9D9D9}87/21}  & \multicolumn{1}{c|}{\cellcolor[HTML]{D9D9D9}120/24} & 51/16   & \multicolumn{1}{c|}{\cellcolor[HTML]{D9D9D9}14/12} & \multicolumn{1}{c|}{\cellcolor[HTML]{D9D9D9}40/10}  & \multicolumn{1}{c|}{\cellcolor[HTML]{D9D9D9}42/8}   & \multicolumn{1}{c|}{\cellcolor[HTML]{D9D9D9}9/8}   & \multicolumn{1}{c|}{\cellcolor[HTML]{D9D9D9}6/8}   & 5/7     \\ \hline
Empty results                                       & \multicolumn{1}{c|}{71/0}                           & \multicolumn{1}{c|}{49/0}                           & \multicolumn{1}{c|}{71/0}                           & 41/0    & \multicolumn{1}{c|}{7/16}                          & \multicolumn{1}{c|}{8/15}                           & \multicolumn{1}{c|}{12/13}                          & \multicolumn{1}{c|}{14/16}                         & \multicolumn{1}{c|}{5/14}                          & 5/14    \\ \hline
Single NULL return                                  & \multicolumn{1}{c|}{13/1}                           & \multicolumn{1}{c|}{13/1}                           & \multicolumn{1}{c|}{14/2}                           & 9/1     & \multicolumn{1}{c|}{0/0}                           & \multicolumn{1}{c|}{0/0}                            & \multicolumn{1}{c|}{0/0}                            & \multicolumn{1}{c|}{2/0}                           & \multicolumn{1}{c|}{2/0}                           & 2/0     \\ \hline
\rowcolor[HTML]{D9D9D9} 
\textbf{SUM: Suspicious Symptoms}                   & \multicolumn{1}{c|}{\cellcolor[HTML]{D9D9D9}84/1}   & \multicolumn{1}{c|}{\cellcolor[HTML]{D9D9D9}62/1}   & \multicolumn{1}{c|}{\cellcolor[HTML]{D9D9D9}85/2}   & 50/1    & \multicolumn{1}{c|}{\cellcolor[HTML]{D9D9D9}7/16}  & \multicolumn{1}{c|}{\cellcolor[HTML]{D9D9D9}8/15}   & \multicolumn{1}{c|}{\cellcolor[HTML]{D9D9D9}12/13}  & \multicolumn{1}{c|}{\cellcolor[HTML]{D9D9D9}16/16} & \multicolumn{1}{c|}{\cellcolor[HTML]{D9D9D9}7/14}  & 7/14    \\ \hline
\rowcolor[HTML]{A6A6A6} 
\textbf{Reported errors}            & \multicolumn{1}{c|}{\cellcolor[HTML]{A6A6A6}481/27} & \multicolumn{1}{c|}{\cellcolor[HTML]{A6A6A6}316/22} & \multicolumn{1}{c|}{\cellcolor[HTML]{A6A6A6}379/27} & 178/25  & \multicolumn{1}{c|}{\cellcolor[HTML]{A6A6A6}55/28} & \multicolumn{1}{c|}{\cellcolor[HTML]{A6A6A6}134/25} & \multicolumn{1}{c|}{\cellcolor[HTML]{A6A6A6}131/21} & \multicolumn{1}{c|}{\cellcolor[HTML]{A6A6A6}57/24} & \multicolumn{1}{c|}{\cellcolor[HTML]{A6A6A6}18/22} & 14/21   \\ \hline \hline \hline
\rowcolor[HTML]{A6A6A6} 
\textbf{Reported erroneous queries} & \multicolumn{1}{c|}{\cellcolor[HTML]{A6A6A6}361/27} & \multicolumn{1}{c|}{\cellcolor[HTML]{A6A6A6}250/23} & \multicolumn{1}{c|}{\cellcolor[HTML]{A6A6A6}311/26} & 157/25  & \multicolumn{1}{c|}{\cellcolor[HTML]{A6A6A6}45/28} & \multicolumn{1}{c|}{\cellcolor[HTML]{A6A6A6}105/31} & \multicolumn{1}{c|}{\cellcolor[HTML]{A6A6A6}105/21} & \multicolumn{1}{c|}{\cellcolor[HTML]{A6A6A6}53/26} & \multicolumn{1}{c|}{\cellcolor[HTML]{A6A6A6}15/22} & 14/36   \\ \hline
\rowcolor[HTML]{A6A6A6} 
\textbf{Actual erroneous queries}   & \multicolumn{1}{c|}{\cellcolor[HTML]{A6A6A6}690}    & \multicolumn{1}{c|}{\cellcolor[HTML]{A6A6A6}598}    & \multicolumn{1}{c|}{\cellcolor[HTML]{A6A6A6}675}    & 502     & \multicolumn{1}{c|}{\cellcolor[HTML]{A6A6A6}238}   & \multicolumn{1}{c|}{\cellcolor[HTML]{A6A6A6}250}    & \multicolumn{1}{c|}{\cellcolor[HTML]{A6A6A6}299}    & \multicolumn{1}{c|}{\cellcolor[HTML]{A6A6A6}191}   & \multicolumn{1}{c|}{\cellcolor[HTML]{A6A6A6}182}   & 146     \\ \hline
\end{tabular}
\begin{tablenotes}
\item [1] The number before slash refers to the detected true error (true positive). The number \edit{after} slash refers to the false reported error (false positive).
\end{tablenotes}
\end{threeparttable}
}
\vspace{-0.1in}
\end{table}

%% file: tables/chapter_5_generalizability_detection.tex
\begin{table}
\centering
\caption{Error Detection results of \tool on additional benchmarks and LLMs \footnotesize{\textnormal{(with MAC-SQL)}}.}
\vspace{-0.1in}
\label{table: generalizability detection}
\resizebox{\linewidth}{!}{%
\begin{threeparttable}
\begin{tabular}{|l|ccccc|ccccc|ccccc|}
\hline
\multicolumn{1}{|c|}{}                              & \multicolumn{5}{c|}{\bird}                                                                                                                                                                                                       & \multicolumn{5}{c|}{\spider}                                                                                                                                                                                                & \multicolumn{5}{c|}{\sciencebenchmark}                                                                                                                                                                                             \\ \cline{2-16} 
\multicolumn{1}{|c|}{\multirow{-2}{*}{}}            & \multicolumn{1}{c|}{GPT3.5}                         & \multicolumn{1}{c|}{Qwen3}                          & \multicolumn{1}{c|}{DSCoder}                        & \multicolumn{1}{c|}{Mistral}                        & LLaMA   & \multicolumn{1}{c|}{GPT3.5}                        & \multicolumn{1}{c|}{Qwen3}                         & \multicolumn{1}{c|}{DSCoder}                       & \multicolumn{1}{c|}{Mistral}                       & LLaMA  & \multicolumn{1}{c|}{GPT3.5}                         & \multicolumn{1}{c|}{Qwen3}                         & \multicolumn{1}{c|}{DSCoder}                        & \multicolumn{1}{c|}{Mistral}                        & LLaMA  \\ \hline
Function Hallucination                              & \multicolumn{1}{c|}{0/0}                            & \multicolumn{1}{c|}{1/0}                            & \multicolumn{1}{c|}{19/0}                           & \multicolumn{1}{c|}{7/0}                            & 41/0    & \multicolumn{1}{c|}{0/0}                           & \multicolumn{1}{c|}{0/0}                           & \multicolumn{1}{c|}{0/0}                           & \multicolumn{1}{c|}{1/0}                           & 0/0    & \multicolumn{1}{c|}{0/0}                            & \multicolumn{1}{c|}{0/0}                           & \multicolumn{1}{c|}{0/0}                            & \multicolumn{1}{c|}{1/0}                            & 0/0    \\ \hline
Other Syntax Violation                              & \multicolumn{1}{c|}{252/0}                          & \multicolumn{1}{c|}{145/0}                          & \multicolumn{1}{c|}{257/0}                          & \multicolumn{1}{c|}{250/0}                          & 423/0   & \multicolumn{1}{c|}{15/0}                          & \multicolumn{1}{c|}{11/0}                          & \multicolumn{1}{c|}{25/0}                          & \multicolumn{1}{c|}{19/0}                          & 139/0  & \multicolumn{1}{c|}{42/0}                           & \multicolumn{1}{c|}{16/0}                          & \multicolumn{1}{c|}{59/0}                           & \multicolumn{1}{c|}{70/0}                           & 217/0  \\ \hline
\rowcolor[HTML]{D9D9D9} 
\textbf{SUM: Syntax Error}                          & \multicolumn{1}{c|}{\cellcolor[HTML]{D9D9D9}252/0}  & \multicolumn{1}{c|}{\cellcolor[HTML]{D9D9D9}146/0}  & \multicolumn{1}{c|}{\cellcolor[HTML]{D9D9D9}276/0}  & \multicolumn{1}{c|}{\cellcolor[HTML]{D9D9D9}257/0}  & 464/0   & \multicolumn{1}{c|}{\cellcolor[HTML]{D9D9D9}15/0}  & \multicolumn{1}{c|}{\cellcolor[HTML]{D9D9D9}11/0}  & \multicolumn{1}{c|}{\cellcolor[HTML]{D9D9D9}25/0}  & \multicolumn{1}{c|}{\cellcolor[HTML]{D9D9D9}20/0}  & 139/0  & \multicolumn{1}{c|}{\cellcolor[HTML]{D9D9D9}42/0}   & \multicolumn{1}{c|}{\cellcolor[HTML]{D9D9D9}16/0}  & \multicolumn{1}{c|}{\cellcolor[HTML]{D9D9D9}59/0}   & \multicolumn{1}{c|}{\cellcolor[HTML]{D9D9D9}71/0}   & 217/0  \\ \hline
Table-Column Mismatch                               & \multicolumn{1}{c|}{126/0}                          & \multicolumn{1}{c|}{17/0}                           & \multicolumn{1}{c|}{45/0}                           & \multicolumn{1}{c|}{62/0}                           & 140/0   & \multicolumn{1}{c|}{4/0}                           & \multicolumn{1}{c|}{0/0}                           & \multicolumn{1}{c|}{14/0}                          & \multicolumn{1}{c|}{8/0}                           & 77/0   & \multicolumn{1}{c|}{27/0}                           & \multicolumn{1}{c|}{8/0}                           & \multicolumn{1}{c|}{49/0}                           & \multicolumn{1}{c|}{82/0}                           & 199/0  \\ \hline
Unused Alias                                        & \multicolumn{1}{c|}{0/0}                            & \multicolumn{1}{c|}{0/0}                            & \multicolumn{1}{c|}{0/0}                            & \multicolumn{1}{c|}{0/0}                            & 0/0     & \multicolumn{1}{c|}{0/0}                           & \multicolumn{1}{c|}{0/0}                           & \multicolumn{1}{c|}{0/0}                           & \multicolumn{1}{c|}{6/0}                           & 6/0    & \multicolumn{1}{c|}{0/0}                            & \multicolumn{1}{c|}{0/0}                           & \multicolumn{1}{c|}{12/0}                           & \multicolumn{1}{c|}{0/0}                            & 12/0   \\ \hline
Schema Hallucination                                & \multicolumn{1}{c|}{16/0}                           & \multicolumn{1}{c|}{7/0}                            & \multicolumn{1}{c|}{7/0}                            & \multicolumn{1}{c|}{9/0}                            & 15/0    & \multicolumn{1}{c|}{0/0}                           & \multicolumn{1}{c|}{0/0}                           & \multicolumn{1}{c|}{0/0}                           & \multicolumn{1}{c|}{0/0}                           & 3/0    & \multicolumn{1}{c|}{22/0}                           & \multicolumn{1}{c|}{4/0}                           & \multicolumn{1}{c|}{6/0}                            & \multicolumn{1}{c|}{16/0}                           & 25/0   \\ \hline
Missing JOIN                                        & \multicolumn{1}{c|}{16/0}                           & \multicolumn{1}{c|}{10/0}                           & \multicolumn{1}{c|}{83/0}                           & \multicolumn{1}{c|}{16/0}                           & 250/0   & \multicolumn{1}{c|}{6/0}                           & \multicolumn{1}{c|}{0/0}                           & \multicolumn{1}{c|}{15/0}                          & \multicolumn{1}{c|}{2/0}                           & 30/0   & \multicolumn{1}{c|}{19/0}                           & \multicolumn{1}{c|}{0/0}                           & \multicolumn{1}{c|}{12/0}                           & \multicolumn{1}{c|}{2/0}                            & 76/0   \\ \hline
\rowcolor[HTML]{D9D9D9} 
\textbf{SUM: Schema Error}                          & \multicolumn{1}{c|}{\cellcolor[HTML]{D9D9D9}158/0}  & \multicolumn{1}{c|}{\cellcolor[HTML]{D9D9D9}34/0}   & \multicolumn{1}{c|}{\cellcolor[HTML]{D9D9D9}135/0}  & \multicolumn{1}{c|}{\cellcolor[HTML]{D9D9D9}87/0}   & 405/0   & \multicolumn{1}{c|}{\cellcolor[HTML]{D9D9D9}10/0}  & \multicolumn{1}{c|}{\cellcolor[HTML]{D9D9D9}0/0}   & \multicolumn{1}{c|}{\cellcolor[HTML]{D9D9D9}29/0}  & \multicolumn{1}{c|}{\cellcolor[HTML]{D9D9D9}16/0}  & 116/0  & \multicolumn{1}{c|}{\cellcolor[HTML]{D9D9D9}68/0}   & \multicolumn{1}{c|}{\cellcolor[HTML]{D9D9D9}12/0}  & \multicolumn{1}{c|}{\cellcolor[HTML]{D9D9D9}79/0}   & \multicolumn{1}{c|}{\cellcolor[HTML]{D9D9D9}100/0}  & 312/0  \\ \hline
Implicit Type Conversion                            & \multicolumn{1}{c|}{4/1}                            & \multicolumn{1}{c|}{4/0}                            & \multicolumn{1}{c|}{17/1}                           & \multicolumn{1}{c|}{10/0}                           & 27/1    & \multicolumn{1}{c|}{0/0}                           & \multicolumn{1}{c|}{0/0}                           & \multicolumn{1}{c|}{0/0}                           & \multicolumn{1}{c|}{0/0}                           & 0/0    & \multicolumn{1}{c|}{0/0}                            & \multicolumn{1}{c|}{0/0}                           & \multicolumn{1}{c|}{0/0}                            & \multicolumn{1}{c|}{0/0}                            & 0/0    \\ \hline
Using = instead of IN                               & \multicolumn{1}{c|}{20/2}                           & \multicolumn{1}{c|}{2/1}                            & \multicolumn{1}{c|}{11/5}                           & \multicolumn{1}{c|}{20/4}                           & 27/9    & \multicolumn{1}{c|}{1/2}                           & \multicolumn{1}{c|}{0/0}                           & \multicolumn{1}{c|}{1/0}                           & \multicolumn{1}{c|}{0/2}                           & 1/0    & \multicolumn{1}{c|}{0/0}                            & \multicolumn{1}{c|}{0/0}                           & \multicolumn{1}{c|}{0/0}                            & \multicolumn{1}{c|}{0/0}                            & 0/0    \\ \hline
Ascending Sort with NULL                            & \multicolumn{1}{c|}{0/0}                            & \multicolumn{1}{c|}{0/1}                            & \multicolumn{1}{c|}{3/0}                            & \multicolumn{1}{c|}{1/0}                            & 1/0     & \multicolumn{1}{c|}{0/0}                           & \multicolumn{1}{c|}{0/0}                           & \multicolumn{1}{c|}{0/0}                           & \multicolumn{1}{c|}{0/0}                           & 0/0    & \multicolumn{1}{c|}{1/0}                            & \multicolumn{1}{c|}{1/0}                           & \multicolumn{1}{c|}{2/0}                            & \multicolumn{1}{c|}{3/0}                            & 3/0    \\ \hline
\rowcolor[HTML]{D9D9D9} 
\textbf{SUM: Logic Error}                           & \multicolumn{1}{c|}{\cellcolor[HTML]{D9D9D9}24/3}   & \multicolumn{1}{c|}{\cellcolor[HTML]{D9D9D9}6/2}    & \multicolumn{1}{c|}{\cellcolor[HTML]{D9D9D9}31/6}   & \multicolumn{1}{c|}{\cellcolor[HTML]{D9D9D9}31/4}   & 55/10   & \multicolumn{1}{c|}{\cellcolor[HTML]{D9D9D9}1/2}   & \multicolumn{1}{c|}{\cellcolor[HTML]{D9D9D9}0/0}   & \multicolumn{1}{c|}{\cellcolor[HTML]{D9D9D9}1/0}   & \multicolumn{1}{c|}{\cellcolor[HTML]{D9D9D9}0/2}   & 1/0    & \multicolumn{1}{c|}{\cellcolor[HTML]{D9D9D9}1/0}    & \multicolumn{1}{c|}{\cellcolor[HTML]{D9D9D9}1/0}   & \multicolumn{1}{c|}{\cellcolor[HTML]{D9D9D9}2/0}    & \multicolumn{1}{c|}{\cellcolor[HTML]{D9D9D9}3/0}    & 3/0    \\ \hline
Violating Value Specification                       & \multicolumn{1}{c|}{49/28}                          & \multicolumn{1}{c|}{79/35}                          & \multicolumn{1}{c|}{90/30}                          & \multicolumn{1}{c|}{74/22}                          & 77/23   & \multicolumn{1}{c|}{21/21}                         & \multicolumn{1}{c|}{13/19}                         & \multicolumn{1}{c|}{19/15}                         & \multicolumn{1}{c|}{23/17}                         & 29/12  & \multicolumn{1}{c|}{75/29}                          & \multicolumn{1}{c|}{44/26}                         & \multicolumn{1}{c|}{84/26}                          & \multicolumn{1}{c|}{75/25}                          & 118/11 \\ \hline
Aggregation/Comparison Misuse                       & \multicolumn{1}{c|}{2/2}                            & \multicolumn{1}{c|}{5/1}                            & \multicolumn{1}{c|}{6/3}                            & \multicolumn{1}{c|}{11/2}                           & 8/4     & \multicolumn{1}{c|}{4/3}                           & \multicolumn{1}{c|}{2/3}                           & \multicolumn{1}{c|}{1/4}                           & \multicolumn{1}{c|}{1/4}                           & 7/3    & \multicolumn{1}{c|}{1/0}                            & \multicolumn{1}{c|}{0/0}                           & \multicolumn{1}{c|}{0/0}                            & \multicolumn{1}{c|}{0/0}                            & 0/1    \\ \hline
Comparing Unrelated Columns                         & \multicolumn{1}{c|}{1/0}                            & \multicolumn{1}{c|}{0/0}                            & \multicolumn{1}{c|}{0/1}                            & \multicolumn{1}{c|}{0/0}                            & 1/0     & \multicolumn{1}{c|}{1/0}                           & \multicolumn{1}{c|}{0/0}                           & \multicolumn{1}{c|}{1/0}                           & \multicolumn{1}{c|}{1/0}                           & 0/0    & \multicolumn{1}{c|}{22/3}                           & \multicolumn{1}{c|}{1/0}                           & \multicolumn{1}{c|}{33/8}                           & \multicolumn{1}{c|}{16/2}                           & 14/1   \\ \hline
\rowcolor[HTML]{D9D9D9} 
\textbf{SUM: Convention Error}                      & \multicolumn{1}{c|}{\cellcolor[HTML]{D9D9D9}52/30}  & \multicolumn{1}{c|}{\cellcolor[HTML]{D9D9D9}84/36}  & \multicolumn{1}{c|}{\cellcolor[HTML]{D9D9D9}96/34}  & \multicolumn{1}{c|}{\cellcolor[HTML]{D9D9D9}85/24}  & 86/27   & \multicolumn{1}{c|}{\cellcolor[HTML]{D9D9D9}26/24} & \multicolumn{1}{c|}{\cellcolor[HTML]{D9D9D9}15/22} & \multicolumn{1}{c|}{\cellcolor[HTML]{D9D9D9}21/19} & \multicolumn{1}{c|}{\cellcolor[HTML]{D9D9D9}25/21} & 36/15  & \multicolumn{1}{c|}{\cellcolor[HTML]{D9D9D9}98/32}  & \multicolumn{1}{c|}{\cellcolor[HTML]{D9D9D9}45/26} & \multicolumn{1}{c|}{\cellcolor[HTML]{D9D9D9}117/34} & \multicolumn{1}{c|}{\cellcolor[HTML]{D9D9D9}91/27}  & 132/13 \\ \hline
Empty results                                       & \multicolumn{1}{c|}{25/4}                           & \multicolumn{1}{c|}{19/5}                           & \multicolumn{1}{c|}{32/3}                           & \multicolumn{1}{c|}{35/4}                           & 52/3    & \multicolumn{1}{c|}{21/10}                         & \multicolumn{1}{c|}{9/10}                          & \multicolumn{1}{c|}{14/11}                         & \multicolumn{1}{c|}{28/10}                         & 24/8   & \multicolumn{1}{c|}{34/5}                           & \multicolumn{1}{c|}{6/5}                           & \multicolumn{1}{c|}{11/6}                           & \multicolumn{1}{c|}{12/6}                           & 26/5   \\ \hline
Single NULL return                                  & \multicolumn{1}{c|}{4/1}                            & \multicolumn{1}{c|}{8/1}                            & \multicolumn{1}{c|}{15/1}                           & \multicolumn{1}{c|}{11/1}                           & 11/0    & \multicolumn{1}{c|}{0/0}                           & \multicolumn{1}{c|}{1/0}                           & \multicolumn{1}{c|}{1/0}                           & \multicolumn{1}{c|}{0/0}                           & 1/0    & \multicolumn{1}{c|}{4/13}                           & \multicolumn{1}{c|}{3/17}                          & \multicolumn{1}{c|}{4/11}                           & \multicolumn{1}{c|}{5/13}                           & 3/4    \\ \hline
\rowcolor[HTML]{D9D9D9} 
\textbf{SUM: Suspicious Symptoms}                   & \multicolumn{1}{c|}{\cellcolor[HTML]{D9D9D9}29/5}   & \multicolumn{1}{c|}{\cellcolor[HTML]{D9D9D9}27/6}   & \multicolumn{1}{c|}{\cellcolor[HTML]{D9D9D9}47/4}   & \multicolumn{1}{c|}{\cellcolor[HTML]{D9D9D9}46/5}   & 63/3    & \multicolumn{1}{c|}{\cellcolor[HTML]{D9D9D9}21/10} & \multicolumn{1}{c|}{\cellcolor[HTML]{D9D9D9}10/10} & \multicolumn{1}{c|}{\cellcolor[HTML]{D9D9D9}15/11} & \multicolumn{1}{c|}{\cellcolor[HTML]{D9D9D9}28/10} & 25/8   & \multicolumn{1}{c|}{\cellcolor[HTML]{D9D9D9}38/18}  & \multicolumn{1}{c|}{\cellcolor[HTML]{D9D9D9}9/22}  & \multicolumn{1}{c|}{\cellcolor[HTML]{D9D9D9}15/17}  & \multicolumn{1}{c|}{\cellcolor[HTML]{D9D9D9}17/19}  & 29/9   \\ \hline
\rowcolor[HTML]{A6A6A6} 
\textbf{Reported errors}            & \multicolumn{1}{c|}{\cellcolor[HTML]{A6A6A6}515/38} & \multicolumn{1}{c|}{\cellcolor[HTML]{A6A6A6}297/44} & \multicolumn{1}{c|}{\cellcolor[HTML]{A6A6A6}585/44} & \multicolumn{1}{c|}{\cellcolor[HTML]{A6A6A6}506/33} & 1073/40 & \multicolumn{1}{c|}{\cellcolor[HTML]{A6A6A6}73/36} & \multicolumn{1}{c|}{\cellcolor[HTML]{A6A6A6}36/32} & \multicolumn{1}{c|}{\cellcolor[HTML]{A6A6A6}91/30} & \multicolumn{1}{c|}{\cellcolor[HTML]{A6A6A6}89/33} & 317/23 & \multicolumn{1}{c|}{\cellcolor[HTML]{A6A6A6}247/50} & \multicolumn{1}{c|}{\cellcolor[HTML]{A6A6A6}83/48} & \multicolumn{1}{c|}{\cellcolor[HTML]{A6A6A6}272/51} & \multicolumn{1}{c|}{\cellcolor[HTML]{A6A6A6}282/46} & 693/22 \\ \hline \hline \hline
\rowcolor[HTML]{A6A6A6} 
\textbf{Reported erroneous queries} & \multicolumn{1}{c|}{\cellcolor[HTML]{A6A6A6}407/38} & \multicolumn{1}{c|}{\cellcolor[HTML]{A6A6A6}268/44} & \multicolumn{1}{c|}{\cellcolor[HTML]{A6A6A6}457/44} & \multicolumn{1}{c|}{\cellcolor[HTML]{A6A6A6}419/33} & 702/40  & \multicolumn{1}{c|}{\cellcolor[HTML]{A6A6A6}66/36} & \multicolumn{1}{c|}{\cellcolor[HTML]{A6A6A6}36/32} & \multicolumn{1}{c|}{\cellcolor[HTML]{A6A6A6}78/30} & \multicolumn{1}{c|}{\cellcolor[HTML]{A6A6A6}76/33} & 224/23 & \multicolumn{1}{c|}{\cellcolor[HTML]{A6A6A6}188/48} & \multicolumn{1}{c|}{\cellcolor[HTML]{A6A6A6}72/48} & \multicolumn{1}{c|}{\cellcolor[HTML]{A6A6A6}202/46} & \multicolumn{1}{c|}{\cellcolor[HTML]{A6A6A6}191/46} & 394/21 \\ \hline
\rowcolor[HTML]{A6A6A6} 
\textbf{Actual erroneous queries}   & \multicolumn{1}{c|}{\cellcolor[HTML]{A6A6A6}743}    & \multicolumn{1}{c|}{\cellcolor[HTML]{A6A6A6}616}    & \multicolumn{1}{c|}{\cellcolor[HTML]{A6A6A6}791}    & \multicolumn{1}{c|}{\cellcolor[HTML]{A6A6A6}765}    & 944     & \multicolumn{1}{c|}{\cellcolor[HTML]{A6A6A6}269}   & \multicolumn{1}{c|}{\cellcolor[HTML]{A6A6A6}255}   & \multicolumn{1}{c|}{\cellcolor[HTML]{A6A6A6}303}   & \multicolumn{1}{c|}{\cellcolor[HTML]{A6A6A6}362}   & 554    & \multicolumn{1}{c|}{\cellcolor[HTML]{A6A6A6}359}    & \multicolumn{1}{c|}{\cellcolor[HTML]{A6A6A6}285}   & \multicolumn{1}{c|}{\cellcolor[HTML]{A6A6A6}352}    & \multicolumn{1}{c|}{\cellcolor[HTML]{A6A6A6}388}    & 485    \\ \hline
\end{tabular}
\begin{tablenotes}
\item [1] The number before slash refers to the detected true error (true positive). The number \edit{after} slash refers to the false reported error (false positive).
\end{tablenotes}
\end{threeparttable}
}
\vspace{-0.15in}
\end{table}

%% file: tables/chapter_5_generalizability_repairing.tex
\begin{table}[h]
\centering
\caption{Error Repairing results of all schemes on additional benchmarks and LLMs \footnotesize{\textnormal{(with MAC-SQL).}}}
\vspace{-0.1in}
\label{table: generalizability repairing}
\resizebox{0.70\linewidth}{!}{%
\begin{threeparttable}
\begin{tabular}{|cc|c|c|c|c|c|c|}
\hline
\multicolumn{2}{|c|}{}                                                    & Rule-Exe & LLM-Plain & LLM-Exe & LLM-Value & LLM-Extr & \textbf{\tool} \\ \hline
\multicolumn{1}{|c|}{\multirow{5}{*}{\bird}}     & GPT-3.5 & 63/1     & 97/41     & 113/38  & 116/48    & 106/56   & \textbf{97/21}                \\ \cline{2-8} 
\multicolumn{1}{|c|}{}                                          & Qwen3   & 29/6     & 47/24     & \textbf{53/21}   & 52/34     & 50/26    & 53/25                \\ \cline{2-8} 
\multicolumn{1}{|c|}{}                                          & DSCoder & 51/0     & 82/26     & 81/23   & \textbf{100/32}    & 83/34    & 78/12                \\ \cline{2-8} 
\multicolumn{1}{|c|}{}                                          & Mistral & 55/2     & 112/52    & 114/51  & 117/56    & 114/62   & \textbf{88/20}                \\ \cline{2-8} 
\multicolumn{1}{|c|}{}                                          & LLaMA   & 70/0     & 57/24     & 98/26   & 109/37    & 62/43    & \textbf{119/16}               \\ \hline
\multicolumn{1}{|c|}{\multirow{5}{*}{\spider}}   & GPT-3.5 & 23/2     & 86/58     & 80/48   & 80/47     & 90/73    & \textbf{68/5}                 \\ \cline{2-8} 
\multicolumn{1}{|c|}{}                                          & Qwen3   & 13/1     & 74/33     & 81/33   & 76/31     & 83/35    & \textbf{65/5}                 \\ \cline{2-8} 
\multicolumn{1}{|c|}{}                                          & DSCoder & 23/0     & 63/38     & 52/41   & 57/45     & 68/45    & \textbf{46/7}                 \\ \cline{2-8} 
\multicolumn{1}{|c|}{}                                          & Mistral & 13/3     & 81/141    & 73/139  & 71/134    & 85/146   & \textbf{59/15}                \\ \cline{2-8} 
\multicolumn{1}{|c|}{}                                          & LLaMA   & 53/1     & 76/69     & 110/60  & 116/70    & 100/127  & \textbf{61/2 }                \\ \hline
\multicolumn{1}{|c|}{\multirow{5}{*}{\scibench}} & GPT-3.5 & 30/6     & 55/38     & 51/25   & 63/34     & 55/39    & \textbf{38/8}                 \\ \cline{2-8} 
\multicolumn{1}{|c|}{}                                          & Qwen3   & \textbf{22/1}     & 35/25     & 38/14   & 37/34     & 35/26    & 19/11                \\ \cline{2-8} 
\multicolumn{1}{|c|}{}                                          & DSCoder & 18/2     & 24/27     & 30/30   & 34/36     & 25/27    & \textbf{34/5}                 \\ \cline{2-8} 
\multicolumn{1}{|c|}{}                                          & Mistral & 35/4     & 57/63     & 59/62   & 54/67     & 58/65    & \textbf{42/9}                 \\ \cline{2-8} 
\multicolumn{1}{|c|}{}                                          & LLaMA   & 25/3     & 9/39      & 32/33   & 33/31     & 9/45     & \textbf{37/3}                 \\ \hline
\end{tabular}
\begin{tablenotes}
\item [1] The number before slash refers to a repair that makes the query correct (true positive). The number \edit{after} slash refers to a repair that makes a correct query wrong (false positive). Bold text refers to the best result of each setting.
\end{tablenotes}
\end{threeparttable}
}
\vspace{-0.15in}
\end{table}

%% file: sections/6-Threats.tex
\noindent\textbf{\textit{Internal Validity.}} 
\edit{Our taxonomy and error labels are based on manual inspection of 4,602 incorrect queries by three co-authors, which may be incorrect.}
\tool assumes that correct SQL queries and meaningful join operations do not return empty result sets or a single data entry with \code{NULL} values, which could be incorrect. 
\edit{We classify SQL queries that lack a clear edit path to correctness into ``unclassifiable'' errors which are excluded from targeted detection and repair.}

\edit{\noindent\textbf{\textit{Construct Validity.}}
We adopt execution match (EM) metric, which may yield false positives when a semantically incorrect query coincidentally produces the same result on a specific database instance. 
This may lead to under-reporting of errors.}

\noindent\textbf{\textit{External Validity.}} 
Our study only cover \bird, \spider and \sciencebenchmark datasets, which may not represent all text-to-SQL tasks.  We only evaluate four ICL-based text-to-SQL techniques incorporated GPT series and 4 open-source LLMs, which may not represent all ICL-based technique designs and LLMs.
\edit{The evaluated techniques are also coupled with their official prompt designs, and different prompt engineering choices may lead to different error distributions and repair behaviors.}

%% file: sections/7-related_work.tex
\noindent \textbf{\textit{Text-to-SQL Techniques.}}
One line of work employs small-scale recurrent networks and variants to generate SQL queries from natural language questions. They either directly generate SQL queries~\cite{PT-MAML,STAMP}, use intermediate representations~\cite{SyntaxSQLNet, IRNet,IncSQL,NatSQL,SmBop}, or fill slot values of a template~\cite{Seq2SQL,SQLNet,TypeSQL,Coarse2Fine}.
Another line of work directly employs\cite{HydraNet,SQLova}, fine-tunes~\cite{Li2023RESDSQL,Ren2024PURPLE,UnifiedSKG,yu2021grappa}, or modifies~\cite{Li2023GraphixT5,BRIDGE,X-SQL} pre-trained language models (PLMs) to achieve higher accuracy.

Recently, LLM-based text-to-SQL techniques have been proposed with the emergence of LLM. Most researchers have adopted in-context learning techniques, including chains of thought~\cite{din-sql,xie-etal-2024-dea}, question decomposition~\cite{wang2024macsql,din-sql,xie-etal-2024-dea}, and self-reflection~\cite{wang2024macsql,din-sql,talaei2024chess,xie-etal-2024-dea}. 
CodeS~\cite{CodeS} and SQL-PaLM~\cite{SQL-PaLM} adopt supervised fine-tuning to enhance the accuracy on certain datasets.

They focus on improving the overall accuracy of SQL query generation. However, they still have many errors to be tackled.

\vspace{3pt}
\noindent \textbf{\textit{SQL Error Analysis.}} 
We have discussed some techniques that analyze the errors of their own generation results in Section~\ref{sec:background_repair}\&\ref{sec:method_effectiveness}. They only observe the consequences of errors and thus have limited error repairing performance. 
Other works~\cite{li2024llm-nl2sql-survey, liu2024surveynl2sqllargelanguage} conduct surveys on text-to-SQL techniques that utilize LLMs.
Ning et al.~\cite{ning2023errorstudy_extend} study errors of PLM-based techniques and summarize the errors based on query clauses. A recent work~\cite{NL2SQL-BUG} studies the text-to-SQL errors in \bird dataset and provide a high-level error taxonomy. However, these work cannot guide automated error detection.
Prior works also study the errors of human-written SQL queries, aiming to assist education~\cite{Ahadi2016StudentsSyntacticMistakes, Taipalus2020ExplainingCausesBehindSQLQueryFormulationErrors,Taipalus2018ErrorsandComplicationsinSQLQueryFormulation,Miedema2022ExpertPerspectivesonStudentErrorsinSQL,Yang2022AnalyzingStudentSQL,Miedema2021IdentifyingSQLMisconceptions}. 
In comparison, we conduct the first comprehensive study of ICL-based text-to-SQL errors and create a taxonomy according to error symptoms, based on which we propose an automatic error repairing framework \tool.

Recent works study metric design to improve evaluation~\cite{zhong-etal-2020-semantic-test-suites, spider, Seq2SQL, li2024llm-nl2sql-survey, kim2024flexexpertlevelfalselessexecution, ascoli2024esmmoderninsightsperspective}. Other works study the ambiguity of NL questions and database schema~\cite{bhaskar-etal-2023-benchmarking, wretblad2024understandingeffectsnoisetexttosql}. They are orthogonal to our study.  

\vspace{3pt}
\noindent \textbf{\textit{Text-to-SQL Error Repairing.}} 
To the best of our knowledge, all existing automatic repairing methods for LLM-generated SQL queries fully rely on LLMs. % to repair errors.
Some works invoke LLMs to repair all generated SQL queries without additional information~\cite{din-sql} or with execution results~\cite{cen2024sqlfixagentsemanticaccuratetexttosqlparsing}. Others only focus on errors with easy-observable consequences like execution failure and empty results~\cite{RSL-SQL,CHASE-SQL,wang2024macsql,MAG-SQL,jain2024contextawaresqlerrorcorrection}, \code{MAX/MIN} operations~\cite{xie-etal-2024-dea}, value specifications~\cite{talaei2024chess,E-SQL,wang2024toolassistedagentsqlinspection}, and other errors~\cite{wang2024dacdecomposedautomationcorrection,li2024seasqlsemanticenhancedtexttosqladaptive}.
As discussed in Section~\ref{sec:method_effectiveness}, they have high computational overhead and many mis-repairs.

Recently, CatSQL~\cite{Fu2023CatSQL} and PURPLE~\cite{Ren2024PURPLE} propose simple rule-based repairing methods for their non-ICL-based techniques.
Chen et al.~\cite{chen2023texttosqlerrorcorrectionlanguage} fine-tunes LLM to repair incorrect SQL queries, assuming perfect error detection.
Yang et al.~\cite{yang2023repairingnaturallanguagesql} propose a human-in-the-loop repairing solution. MAGIC~\cite{askari2024magicgeneratingselfcorrectionguideline} provides fixing guidelines that require human reviews. 

In comparison, our \tool achieves low overhead and neglectable mis-repairs, with its rule-based solutions based on our in-depth understanding of text-to-SQL errors. 

%% file: sections/8-conclusion.tex
In this paper, we have conducted the first comprehensive study on ICL-based text-to-SQL errors and the effectiveness of existing automated repairing methods. 
We find that text-to-SQL errors are widespread and have summarized 27 error types of 7 categories. 
We also find that existing methods have small correctness improvement, as LLMs have limited self-correcting capability.
Based on the study, we have proposed \tool, an automatic framework that efficiently detects and repairs SQL queries generated by ICL-based techniques. It improves the overall correctness of generated SQL queries while having minimal mis-repairs and computation overhead.
We hope that our study and \tool can inspire future research in improving automated code generation and ICL-based text-to-SQL techniques.